\documentclass[runningheads]{llncs}

\usepackage[mobile]{eccv}

\usepackage{eccvabbrv}

\usepackage{graphicx}
\usepackage{booktabs}
\usepackage[ruled, lined, linesnumbered, commentsnumbered, longend]{algorithm2e}

\usepackage{tikz}
\usepackage{pgfplots}
\usepgfplotslibrary{groupplots}
\usepackage{siunitx}
\usepackage{comment}
\usepackage{xfrac}
\usepackage{bm}
\usepackage{multirow}
\usepackage[dvipsnames]{xcolor}

\usepackage{hyperref}

\usepackage{orcidlink}

\begin{document}

\title{LEROjD:  \underline{L}idar \underline{E}xtended \underline{R}adar-Only \underline{O}b\underline{j}ect \underline{D}etection} 


\author{Patrick Palmer\inst{1}\orcidlink{0000-0003-0223-1112} \and
Martin Krüger\inst{1}\orcidlink{0000-0003-0544-0331} \and
Stefan Schütte\inst{1}\orcidlink{0000-0003-3126-4626} \and
Richard Altendorfer\inst{2}\orcidlink{0000-0003-2884-8444} \and 
Ganesh Adam\inst{2}\orcidlink{0000-0003-1851-5611} \and 
Torsten Bertram\inst{1}\orcidlink{0000-0002-6096-8190}}

\authorrunning{P.~Palmer et al.}

\institute{Institute of Control Theory and Systems Engineering, TU Dortmund University
\email{\{patrick.palmer, martin2.krueger, stefan.schuette, torsten.bertram\}@tu-dortmund.de}\\
\and
ZF Group,
\email{\{richard.altendorfer, ganesh.adam\}@zf.com}}
\maketitle

\begin{abstract}
  Accurate 3D object detection is vital for automated driving. While lidar sensors are well suited for this task, they are expensive and have limitations in adverse weather conditions. 3+1D imaging radar sensors offer a cost-effective, robust alternative but face challenges due to their low resolution and high measurement noise. Existing 3+1D imaging radar datasets include radar and lidar data, enabling cross-modal model improvements. Although lidar should not be used during inference, it can aid the training of radar-only object detectors. We explore two strategies to transfer knowledge from the lidar to the radar domain and radar-only object detectors: 1. multi-stage training with sequential lidar point cloud thin-out, and 2. cross-modal knowledge distillation. In the multi-stage process, three thin-out methods are examined. Our results show significant performance gains of up to 4.2 percentage points in mean Average Precision with multi-stage training and up to 3.9 percentage points with knowledge distillation by initializing the student with the teacher's weights. The main benefit of these approaches is their applicability to other 3D object detection networks without altering their architecture, as we show by analyzing it on two different object detectors. Our code is available at \url{https://github.com/rst-tu-dortmund/lerojd}.
  \keywords{3D Object Detection \and 3+1D Imaging Radar \and Cross-Modal Object Detection}
\end{abstract}

\section{Introduction}
\label{sec:intro}
Environment perception is the first module in each automated driving stack. Multiple sensor modalities, like cameras, lidars, and radars are utilized for this task. Radar sensors are of unique interest in perception due to their robustness against poor lighting, challenging weather conditions like rain or snow, and cost-effectiveness \cite{Brisken2018RecentRadarEvaluation, Sun20214DAutomotiveRadarSensingSparsity, palffy2022voddataset}. One exclusive advantage is the ability to measure the relative radial velocity of reflections directly due to the Doppler effect.

While a precise localization of objects is possible with traditional radar sensors without elevation angle measurements \cite{Danzer20192DRadarCarDetection}, it is inherently limited to the horizontal plane. Additionally, it is hard to predict the objects' extent due to the reflections' sparsity. The introduction of 3+1D high-resolution imaging radar sensors has recently mitigated these limitations, at least partially. In addition to measuring the elevation angle of reflections, the density of measurements is increased \cite{Jiang20234DHighResolutionImagery, Engels2021AutomotivRadarSignalProcessing}. Therefore, approaches that only utilize radar sensors for perception of the environment are of particular interest.

Despite improvements in radar-based object detection, the performance still lags behind other sensor modalities like lidar \cite{palffy2022voddataset}. One persistent major limitation of 2+1D classic radar and 3+1D imaging radar sensors is the relative sparsity of the point cloud, which limits the detection performance. 

Lidar sensors, on the other hand, are well suited for object detection and are therefore frequently employed as a reference for evaluating the performance of different sensor modalities due to their ability to produce an accurate and dense understanding of the scene. Their effectiveness is particularly pronounced in detecting nearby traffic participants without occlusion \cite{Wu2023VirtualSparseConvolution}.

All currently available datasets that contain 3+1D imaging radar data additionally accommodate lidar sensor data \cite{palffy2022voddataset, Rebut2022radial, Zheng2022tj4drad, paek2022kradar, meyer2019astyx, Choi2023MSCRadar, caesar2020nuscenes, zhang2023dualradar, ZhangZhugeLiu2023ITSC}. The lidar sensor data of these datasets is currently either utilized for labeling, combining multiple sensor modalities for accurate object detection, or comparing the performance of radar-only techniques to another sensor modality. While the majority of series production vehicles may not include lidar sensors due to cost and vehicle design constraints, they will still be available in the training process of learning-based methods. Extending radar-only methods with lidar sensor data in training has been shown to be a viable method for estimating point flow on the imaging radar point cloud \cite{Ding2023HiddenGems}. These observations lead to the following research question: \textit{Can lidar sensor data be utilized in the training process of imaging radar-based 3D object detectors to improve the object detection performance on radar-only data during inference?}

\begin{figure}[t]
  \centering
\resizebox{11.5cm}{!}{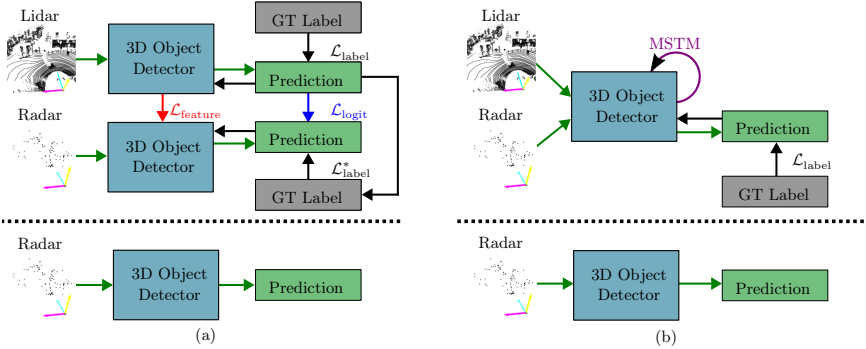}
  \caption{Architecture overview of (a) a knowledge distillation-based method and (b) a multi-stage training method (MSTM) for utilizing lidar data in the training of radar-only object detectors. The ground truth (GT) label is the same for both methods. The diagrams above the dotted line represent the training process, while the diagrams below the dotted line represent inference.}
  \label{fig:FirstPage}
\end{figure}

To use different sensor modalities in training, transfer learning and knowledge distillation (KD) principles can be utilized. While KD is commonly employed across sensor modalities such as camera images and lidar point clouds, its application between 3+1D imaging radar and lidar sensors remains unexplored. Since lidar and imaging radar share a structurally similar data representation as point clouds, an identical base network with different input modalities can be utilized to transfer knowledge between different input modalities. We investigate two approaches to transfer knowledge from lidar-based to radar-based object detectors: a KD-based and a multi-stage training approach.
The main principles of the two methods are visualized in Figure \ref{fig:FirstPage}.

The contributions of this work are summarized as follows:
\begin{itemize}
  \item We investigate a combination of lidar and radar sensors in the training stage of object detectors to improve radar-only object detection at inference.
  \item We investigate three thin-out strategies for lidar point clouds to transfer knowledge from dense lidar to sparse lidar and radar-only object detectors. 
  \item We propose a multi-stage training procedure to transfer knowledge from dense lidar to sparse lidar and, finally, to radar-only object detection.
  \item We modify and analyze several knowledge distillation-based approaches to transfer knowledge from lidar to radar-only object detectors.
\end{itemize}
\section{Related Work}
\label{sec:related}

Imaging radar sensors commonly utilize the point cloud as a data representation format instead of the radar tensor, due to its higher computational efficiency. This format is similar to the one used for lidars, enabling the application of object detection methods developed for lidars to radars. Object detection on lidars can be split into two main categories. Point-based methods, like PointNet++ \cite{qi2017pointnet},  downsample the original point cloud, encode it using a backbone, and finally apply a detection head. Voxel-based methods discretize the point cloud into a 3D grid and apply 3D convolutions to the grid before finally applying a detection head \cite{Chen2023VoxelNext, zhou2018voxelnet, deng2021voxelrcnn}. The main drawback of voxel-based methods is the high memory consumption and the loss of spatial information due to the discretization of the point cloud. To overcome the limitation of high memory consumption, \cite{lang2019pointpillars, Li2023PillarNext} have proposed the PointPillars network. Pillars are a specific form of voxels that span over the full height of the scene and represent the point cloud in a 2D grid. 

Methods from the lidar domain have been shown to perform reasonably well on radar data \cite{Palmer2023Reviewing3DObjectDetectors, palffy2022voddataset}. However, they are limited due to the sparsity of the radar point cloud. SMURF \cite{Liu2023SMURF} considers two representations of the radar point cloud to address sparsity. Using kernel density estimation, it utilizes pillarization and density features derived from a multi-dimensional Gaussian mixture distribution. RPFA-Net \cite{Xu2021RPFA} is a PointPillars \cite{lang2019pointpillars} based network, which introduces a self-attention mechanism to extract global context information from the radar point cloud. RadarMFNet \cite{tan20223d} utilizes a multi-frame radar point cloud representation to address the sparsity of the radar point cloud in conjunction with an anchor-based detector and temporal pooling layers. 

One way of improving 3D object detection on radar data is the fusion with additional sensor modalities like camera \cite{Zhou2023RadarCameraFusionViewDisparity, Xiong2023LXL, Zheng2023RCFusion}, lidar \cite{Wang2023MultiModalMultiScaleFusion} or camera and lidar \cite{drews2022deepfusion, chen2023futr3d, Xiao2023DeepFusionLRC} at the cost of introducing additional sensors at inference. 

The concept of knowledge distillation was first introduced by \cite{hinton2015distilling}. It consists of two networks, which are labeled as teacher and student. The teacher is a large and complex model, while the student is less complex and more computationally efficient. The student is trained to mimic the teacher network's performance by utilizing the teacher's predictions and the ground truth labels. This has been utilized in the context of 3D object detection by \cite{Yang2022KDOpenPCDet, Zhang2023PointDistiller, Li2023Lightweigt3DPointCloud} to construct computation time-efficient models that have similar performance as large models. \cite{Li2023LidarGuided} and \cite{Klingner2023X3KD} have extended upon this concept by utilizing KD to extract knowledge from a teacher trained on lidar data to a student who utilizes camera images.

To the best of our knowledge, few studies have investigated the effect of KD considering radar data. A transfer of knowledge from an image-based teacher network to a radar-based student has been shown to improve the performance for the task of people counting \cite{Michael2021PeopleCounting}. For the task of 3D object detection, \cite{Klingner2023X3KD} has shown that a transfer learning-based approach from a lidar and image-based network to the radar domain results in improved object detection performance for classical 2+1D radar sensors. The main drawback of \cite{Klingner2023X3KD} is that in addition to 3D object labels, instance segmentation labels for the image domain are required. Additionally, \cite{Klingner2023X3KD} introduces multiple sub-networks to derive the KD losses, which makes the model more complex, while we aim for a simpler approach. HiddenGems \cite{Ding2023HiddenGems} utilizes lidar point clouds to derive point flow information and train a network to predict the point flow on radar point clouds. See Beyond Seeing \cite{deng2023see} utilizes lidar point cloud for feature hallucination on radar point clouds. Both methods require the lidar point cloud only in the training process but require major modifications to the network architecture.

Sampling of point clouds is a broadly explored topic to reduce the complexity of computing large point clouds. Most commonly, farthest point sampling (FPS) \cite{yang20203dssd, qi2017pointnet, Shi2019PointRCNN}, voxelized FPS \cite{shi2020pv} or random sampling \cite{zhou2018voxelnet} are utilized in the context of 3D object detection. Random sampling has been shown to be beneficial when used for semantic segmentation of point clouds \cite{hu2020randla} and to outperform farthest point sampling at this task \cite{li2023comparative}. One limitation of random sampling is that while points at a close distance are kept, points at a far distance are more likely to be removed. Therefore, \cite{cheng2022pcb} proposes a polar cylinder balanced random sampling to keep a more balanced distribution of points across the range.

\section{Method}
\label{sec:method}
Two methods of transferring knowledge from lidar-based to radar-based object detectors are investigated: Knowledge Distillation (KD) (Section \ref{sec:KD_Methods}), which is modified for the task of transfer learning and our proposed multi-stage training procedure with sequential point cloud thin-out (Section \ref{sec:MS_Methods}). Additionally, the utilized thin-out strategies for lidar point cloud (Section \ref{sec:thin_out}) are described.

\subsection{Knowledge Distillation}
\label{sec:KD_Methods}
KD is commonly used for two tasks. First, designing computationally efficient models by transferring knowledge from a larger teacher network to a smaller student network \cite{tung2019similarity, hinton2015distilling, wang2019distilling}. Second, to transfer knowledge across sensor modalities by utilizing different model architectures for the teacher and student \cite{gupta2016cross, zhao2018through, afouras2020asr}. Lidar and radar-based point cloud object detection can utilize the same model structure but with different input modalities. This enables the utilization of KD methods first described for designing computationally efficient models for cross-modal knowledge transfer from lidar- to radar-based object detectors. In this case, the teacher is trained on the full lidar point cloud, while the student is trained on the radar point cloud. Three different loss terms, as described by \cite{Yang2022KDOpenPCDet}, are employed:

\textbf{Logit-KD} is the first, classic type of distilling knowledge described by \cite{hinton2015distilling}. For 3D object detection, the logit-KD loss $\color{blue}\mathcal{L}_{\text{logit}}$ is split into a classification $\mathcal{L}_{\text{l-cls}}$ and regression loss $\mathcal{L}_{\text{l-reg}}$. These losses are calculated by comparing the student's and teacher's predictions utilizing the 3d detectors' regression loss and bi-linear interpolation between student and teacher output classes.

\textbf{Feature-KD} is widely utilized in 2D object detection \cite{li2017mimicking, wang2019distilling}. It utilizes a loss term that forces the student network to mimic the teacher's intermediate feature map (feat). A feature mimicking the last layer of the bird's eye view feature encoder, similar to \cite{Yang2022KDOpenPCDet}, is utilized in this work.

\textbf{Label-KD} is a recent distillation approach that leans on the concept of the Logit KD but simplifies and generalizes it. It is first described by \cite{Nguyen2022ImprobingODbyLA}. The teacher predictions are filtered by their scores using a score threshold, and an adapted ground truth set is constructed by combining the filtered predictions and the ground truth set. This adapted set is utilized in student training. The loss is split into a classification $\mathcal{L}_{\text{cls}}$ and regression loss $\mathcal{L}_{\text{reg}}$. It replaces $\color{black}\mathcal{L}_{\text{label}}$ usually calculated on the ground truth set.

The three KD loss terms are combined into a joint loss weighted with $\lambda$
\begin{equation}
	\label{eq:loss}
	\mathcal{L_{\text{joint}}} = 
	\underbrace{\lambda_{\text{l-reg}} \mathcal{L}_{\text{l-reg}} + \lambda_{\text{l-cls}} \mathcal{L}_{\text{l-cls.}}}_{\color{blue}\mathcal{L}_{\text{logit}}} + 
	\underbrace{\lambda_{\text{feat}} \mathcal{L}_{\text{feat}}}_{\color{red}\mathcal{L}_{\text{feature}}} +
	\underbrace{\lambda_{\text{reg}} \mathcal{L}_{\text{reg}} + \lambda_{\text{cls}} \mathcal{L}_{\text{cls}}}_{\color{black}\mathcal{L}^{*}_{\text{label}}}
\end{equation}

\subsection{Multi-Stage Lidar Thin-Out Training Procedure}
\label{sec:MS_Methods}
\begin{figure*}[t]
	\centering
	\resizebox{11.7cm}{!}{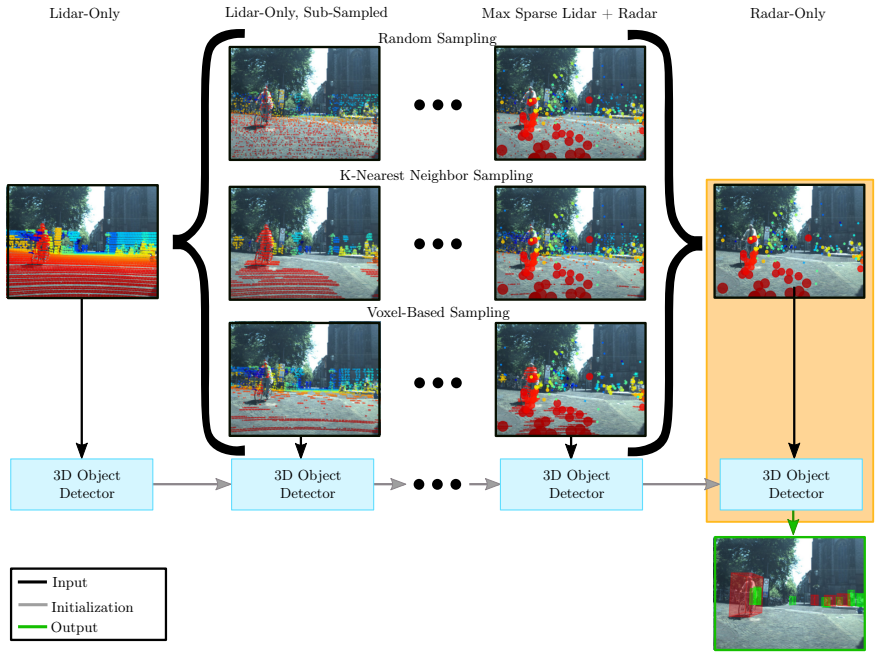}
	\caption{MSTM pipeline. A 3D object detection network is iteratively trained on increasingly sparse lidar data. Three different thin-out strategies are utilized for sparsification. The lidar point cloud is mixed with the radar point cloud in the second to last step. In the last step, the network is only trained on radar data. At inference, only the orange-shaded part is executed. Small points represent lidar points, large points radar points. The color of points corresponds to the distance from the ego-vehicle. The camera image is not used as an input but only aids the visualization.}	 
	\label{fig:ms_archiecture}
\end{figure*}

Utilizing a different input data modality for either pre-training a network on a large dataset or utilizing simulated data in the training process of a point cloud-based network has been shown to improve the object detection performance \cite{Xiao2022transfer}. The multi-stage training method (MSTM) proposed in this work extends upon this by utilizing a Curriculum learning \cite{bengio2009curriculum} based training procedure by which the network is trained on iteratively sparsified lidar point clouds, similar to \cite{wei2022lidar}, and fine-tuned on the radar point cloud.
Figure \ref{fig:ms_archiecture} visualizes our multi-stage training procedure. The network is first trained on the full lidar point cloud until convergence. In the following steps, the lidar point cloud is thinned-out iteratively by a factor of 2 and utilized for training a network whose weights are initialized using the previously trained model. This forces the network to learn features for a good object detection performance on increasingly sparser point clouds. In the second to last step, the lidar point cloud is mixed with the radar points so that the network can translate from the lidar to the radar domain. In the last step, the model is trained only on radar points. A training without multiple stages is called single-stage training method (SSTM) in this work.

In addition to the training where only lidar points are utilized in the first training stages, we investigate the utilization of the radar point cloud in conjunction with the lidar point cloud in all stages. The thin-out of the lidar points remains the same and is mixed with the radar point cloud in each step. This conditions the model on radar from the first step in order to prioritize features in the lidar point cloud that relate to a good object detection on radar-only data.

For mixing lidar and radar point clouds, the voxel or pillar feature encoder is modified to prioritize the radar point cloud in the random sampling process as done by \cite{Nobis2021RadarVoxel}. Otherwise, the vastly higher number of lidar points, even when thinned-out, might lead to the complete exclusion of radar points.

\subsection{Lidar Thin-Out Strategies}
\label{sec:thin_out}
Three different methods for sub-sampling the lidar point cloud are investigated. Figure \ref{fig:ms_archiecture} shows examples for each thin-out stage.

\textbf{Random sampling}
is the simplest method of sparsifying a point cloud. It neglects the structure and inherent limitations of the point cloud representation, especially for objects far away or with a high degree of occlusion. Neglecting the structure can lead to the complete loss of information for objects represented only by few points.

\textbf{K-nearest neighbor sampling}
approximates the reflection density distribution of the radar point cloud with the lidar point cloud by only keeping lidar reflections close to radar reflections. This algorithm is described in Algorithm \ref{alg:kne}. Objects not detected by the radar sensor are, therefore, also not represented by the k-nearest neighbor thinned-out lidar point cloud.

\textbf{Voxel-based sampling}
 aims to reduce the number of points in each area of the point cloud while keeping the general distribution of the point cloud. This is motivated by the fact that radar sensors do not suffer as much from loss of resolution with distance as lidar sensors. The approach is described in Algorithm \ref{alg:vox}, and executed iteratively for a sequence of sparsification steps. 

\begin{algorithm}[H]
	\SetKwFunction{isOddNumber}{isOddNumber}
	\SetKwInOut{KwIn}{Input}
	\SetKwInOut{KwOut}{Output}
	\KwIn{Lidar point cloud $\textit{\textbf{L}} \in \mathbb{R}^{N \times 3}$, radar point cloud $\textit{\textbf{R}} \in \mathbb{R}^{M \times 3}$}
	\KwOut{Sub-sampled lidar point cloud $\textit{\textbf{L}}_s \in \mathbb{R}^{K \times 3}$}
	Calculate the euclidean distance $c_i$ between each lidar point $\textit{\textbf{l}}_i \in \textit{\textbf{L}}$ and its nearest radar point from $\textit{\textbf{R}}$\;
	Select a share of $K$ lidar points with the smallest distances $\textit{\textbf{c}}$\;
	\textbf{return} Share of Points $\textit{\textbf{L}}_s \in \mathbb{R}^{K \times 3}$
	\caption{K-nearest neighbor sampling} \label{alg:kne}
\end{algorithm}
\begin{algorithm}[H]
	\SetKwFunction{isOddNumber}{isOddNumber}
	\SetKwInOut{KwIn}{Input}
	\SetKwInOut{KwOut}{Output}
	
	\KwIn{Lidar point cloud $\textit{\textbf{L}} \in \mathbb{R}^{N \times 3}$}
	\KwOut{Sub-sampled lidar point cloud $\textit{\textbf{L}}_s \in \mathbb{R}^{K \times 3}$}
	
	Initialize $\textit{\textbf{L}}_s$ with $\textit{\textbf{L}}$\;
	Voxelize the lidar point cloud into $v$ voxels\;
	Calculate the number of points in each voxel $\bm{P}_v$\;
	Calculate the minimum number of points per voxel $p_\text{min}$, so that at least $0.75 \ N$ points are in voxels with more than $p_\text{min}$ points\;
	From each voxel with more than  $p_\text{min}$ points, choose  $p_\text{min}$ random points to keep and add remaining points to $\textit{\textbf{L}}_p$\;
	Randomly select $0.5 \ N$ points from $\textit{\textbf{L}}_p$ and remove them from $\textit{\textbf{L}}_s$\;
	\textbf{return} Share of Points $\textit{\textbf{L}}_s \in \mathbb{R}^{K \times 3}$
	\caption{Voxel-based sampling} \label{alg:vox}
\end{algorithm}

Another common sparsification method utilized for imitating low-resolution lidar sensors is layer-based sampling \cite{you2020pseudo, wei2022lidar}. This approach is not investigated because radar sensors do not capture the environment on a layer basis.

\section{Experimental Evaluation}
\label{sec:experiments}
\subsection{Experimental Setup}

\newcommand{\veryshortarrow}[1][3pt]{\mathrel{%
		\hbox{\rule[\dimexpr\fontdimen22\textfont2-.2pt\relax]{#1}{.4pt}}%
		\mkern-4mu\hbox{\usefont{U}{lasy}{m}{n}\symbol{41}}}}

\textbf{Dataset:}
All experiments are conducted utilizing the \textit{View-of-Delft} (VoD) dataset \cite{palffy2022voddataset}. It contains synchronized data of multiple sensor modalities. The 64-layer lidar sensor and the imaging radar are utilized in this work. A point cloud accumulated over 5 frames \cite{palffy2022voddataset}, which has been shown to improve object detection performance compared to no accumulation, is used for radar data \cite{Palmer2023EgoMotionDynamicMotion, palffy2022voddataset}. We detected a duplication of identical points in the lidar point cloud, which could adversely affect all sampling methods; thus, we eliminated the duplicated points from the point cloud. Although the VoD dataset is among the best currently available datasets for imaging radar-based object detection, it is limited by its size compared to other automotive datasets without imaging-radar, like \cite{Sun_2020_CVPR, mao2021one}. 
Given the absence of publicly available labels and limited online evaluation for the test dataset, we repurpose the validation dataset as a test set. Consequently, we partition the original training set into a new training set (\SI{80}{\percent}) and a dedicated validation set (\SI{20}{\percent}) to ensure robust model training. 

\textbf{Evaluation Metrics:}
The primary performance metric utilized to compare the results is the mean average precision (mAP), as used by \cite{Geiger2012KITTI}\cite{Palmer2023Reviewing3DObjectDetectors}. Similar to the evaluation of the Waymo data set \cite{sun2020scalability} in \cite{zamanakos2021comprehensive}, we split the results into two distance bins: short-range (SR): \SI{0}{}-\SI{30}{\meter} and mid-range (MR): \SI{30}{}-\SI{50}{\meter}. All experiments are conducted utilizing three different random network initializations that are averaged.

\textbf{Training:}
Most experiments use the PointPillars model \cite{lang2019pointpillars} as an object detector with the same configuration as utilized by \cite{palffy2022voddataset}. For imaging radar data, PointPillars has been shown to perform among the best out of multiple state-of-the-art 3D object detection methods while still performing adequately on lidar data \cite{Palmer2023Reviewing3DObjectDetectors}. Furthermore PointPillars is a relevant baseline for radar-specific object detection methods \cite{Xu2021RPFA, yan2023mvfan}. To show that the proposed MSTM and KD apply to various object detectors, the most promising approaches from the evaluations on PointPillars are evaluated on DSVT-P \cite{wang2023dsvt}, as an example for a transformer-based model. All SSTM trainings are conducted with an early stopping policy for a maximum of 125 epochs. For the MSTM the initial training on the full lidar point cloud is conducted for 125 epochs, while each refinement step is trained for 30 epochs. All trainings utilize the Adam optimizer \cite{kingma2014adam} and an adapted learning rate scheduler that reaches its maximum learning rate earlier and has a faster descent than the scheduler described by \cite{smith2019super}. This improves the object detection performance on radar data. 

\textbf{Notation:}
To distinguish between methods, the following notation is used:

\begin{align*}
{\color{teal}\mathcal{T\,}}^{{\color{orange}\mathcal{TM}}}
_{{\color{violet}\mathcal{LS}} / {\color{cyan}\mathcal{TO}}} \;\; \rightarrow {\color{teal}\mathcal{T\,}}^{\color{purple}\mathcal{KD}}.
\end{align*}

The notation is split into two parts. The part left of the arrow represents the data set utilized for pre-training, while the right part represents the data and training method utilized for the last (fine-tuning) training stage of the model. This part is omitted if it is the same as the training dataset. The training data $\color{teal}\mathcal{T}$ can either be lidar (L), radar (R), or mixed radar + lidar data (RL). The training method $\color{orange}\mathcal{TM}$ can denote either MSTM or SSTM. The lidar share $\color{violet}\mathcal{LS}$ is the fraction of the original lidar point cloud that is utilized. Radar-only training always uses the full radar point cloud, therefore  $\color{violet}\mathcal{LS}$ is omitted. For the MSTM, this is represented by a range of fractions that are iterated in training. The thin-out method $\color{cyan}\mathcal{TO}$ can either be random (rand), k-nearest neighbor (knn) or voxel-based (vox). $\color{purple}\mathcal{KD}$ represents the KD method; this can either be label (lab), logit (log), feature (feat), or a combination of those (joint). $\color{purple}\mathcal{KD}$ is omitted if just an initialization and fine-tuning is utilized. Examples for the training corresponding to specific notations are listed in Table \ref{tab:notation}.

\begin{table*}[t]
	\centering
	\caption{Examples of training steps and evaluation data corresponding to the notation, empolying voxel-based sub-sampling. Voxel sampling is shortened to v for legibility.}
	\begin{tabular}{l | l | l}
		\toprule   
		Notation & Training & Evaluation\\
		\midrule
		$\text{L}^{\text{MSTM}}_{\text{1-\sfrac{1}{8}/v}} $ & 
		$\text{L}_\text{1/v}$ $\;\;\,\;\, \veryshortarrow$ 
		$\text{L}_\text{\sfrac{1}{2}/v}$ $\;\;\, \veryshortarrow$ 
		$\text{L}_\text{\sfrac{1}{4}/v}$ $\;\;\, \veryshortarrow$ 
		$\text{L}_\text{\sfrac{1}{8}/v}$  & $\text{L}_\text{\sfrac{1}{8}/v}$ \\
		
		$\text{RL}^{\text{MSTM}}_{\text{1-\sfrac{1}{8}/v}} $ &
		$\text{RL}_\text{1/v}$ $\;\,\veryshortarrow$
		$\text{RL}_\text{\sfrac{1}{2}/v}$ $\veryshortarrow$ 
		$\text{RL}_\text{\sfrac{1}{4}/v}$ $\veryshortarrow$ 
		$\text{RL}_\text{\sfrac{1}{8}/v}$ & $\text{RL}_\text{\sfrac{1}{8}/v}$ \\
		\midrule
		$\text{L}^{\text{MSTM}}_{\text{1-\sfrac{1}{16}/v}}\rightarrow\text{R} $ & 
		$\text{L}_\text{1/v}$ $\;\;\,\;\, \veryshortarrow$ 
		$\text{L}_\text{\sfrac{1}{2}/v}$ $\;\;\, \veryshortarrow$
		$\text{L}_\text{\sfrac{1}{4}/v}$ $\;\;\, \veryshortarrow$
		$\text{L}_\text{\sfrac{1}{8}/v}$ $\;\;\, \veryshortarrow$ 
		$\text{RL}_\text{\sfrac{1}{16}/v}$ $\veryshortarrow$ 
		$\text{R}$ & $\text{R}$ \\
		
		$\text{RL}^{\text{MSTM}}_{\text{1-\sfrac{1}{16}/v}}\rightarrow\text{R} \;\;\, $ & 
		$\text{RL}_\text{1/v}$ $\;\,\veryshortarrow$ 
		$\text{RL}_\text{\sfrac{1}{2}/v}$ $\veryshortarrow$ 
		$\text{RL}_\text{\sfrac{1}{4}/v}$ $\veryshortarrow$ 
		$\text{RL}_\text{\sfrac{1}{8}/v}$ $\veryshortarrow$ 
		$\text{RL}_\text{\sfrac{1}{16}/v}$ $\veryshortarrow$ 
		$\text{R} \;\;\;\, $ & $\text{R}$ \\
		\midrule
		$\text{RL}^{\text{SSTM}}_{\text{\sfrac{1}{4}/v}}\rightarrow\text{R} $ & 
		$\text{RL}_{\text{\sfrac{1}{4}/v}}$ $\veryshortarrow$ 
		$\text{R}$ & $\text{R}$ \\
		\bottomrule
	\end{tabular}
	\label{tab:notation}
\end{table*}

\subsection{Evalutation of MSTM on Lidar-only and Mixed Radar + Lidar} \label{sec:MultiStageLidar}
To evaluate the applicability of our proposed MSTM to the thinned-out lidar point cloud, we evaluate the training without the last two steps involving radar data. Training is conducted for  just the lidar point cloud and the mixed radar + lidar point cloud in all stages. The multi-stage trained network is evaluated after each thin-out stage on the thinned-out lidar (or mixed radar + lidar) point cloud. Thin-out stages up to $\sfrac{1}{256}$ of the original lidar point cloud are considered due to the lidar point cloud containing fewer points than the radar point cloud at $\sfrac{1}{256}$ of the original lidar point cloud. The MSTM is compared to the SSTM, which is trained only on the thinned-out point cloud. The results are shown in Fig. \ref{fig:MultistageLidarRadar}, complete quantitative results are given in the supplementary.

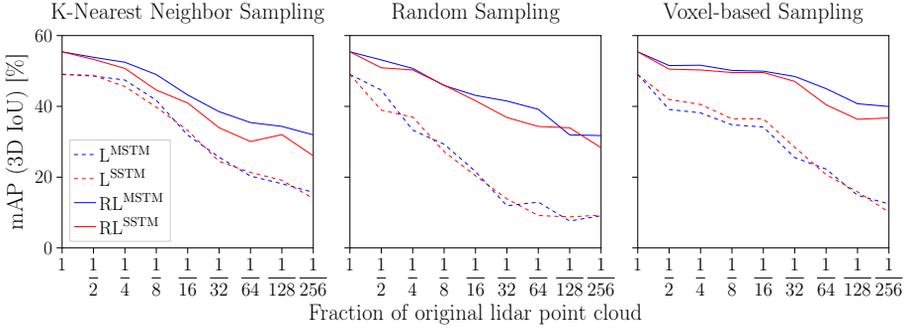
\begin{figure*}[t!]
	\centering
	\begin{tikzpicture}[scale=0.49, yscale=1]
	
	\definecolor{darkgray176}{RGB}{176,176,176}
	\definecolor{lightgray204}{RGB}{204,204,204}
	
	\begin{groupplot}[group style={group size=3 by 1}]		
	
	\nextgroupplot[
	legend cell align={left},
	legend style={
		fill opacity=0.8,
		draw opacity=1,
		text opacity=1,
		at={(0.03,0.03)},
		anchor=south west,
		draw=lightgray204
	},
	tick align=outside,
	tick pos=left,
	title={\LARGE K-Nearest Neighbor Sampling},
	x grid style={darkgray176},
	xmin=0, xmax=8,
	xtick style={color=black},
	xtick={0,1,2,3,4,5,6,7,8},
	xticklabels={
		\Large 1,
		\Large \(\displaystyle \dfrac{1}{2}\),
		\Large \(\displaystyle \dfrac{1}{4}\),
		\Large \(\displaystyle \dfrac{1}{8}\),
		\Large \(\displaystyle \dfrac{1}{16}\),
		\Large \(\displaystyle \dfrac{1}{32}\),
		\Large \(\displaystyle \dfrac{1}{64}\),
		\Large \(\displaystyle \dfrac{1}{128}\),
		\Large \(\displaystyle \dfrac{1}{256}\)
	},
	y grid style={darkgray176},
	ymin=0, ymax=60,
	ytick style={color=black},
	ylabel={\LARGE mAP (3D IoU) [\%]},
	yticklabels={			
		\Large 0,
		\Large 0,
		\Large 20,
		\Large 40,
		\Large 60},
	]
	\addplot [semithick, blue, dashed]
	table {%
		0 49.0152341518965
		1 48.5626125418712
		2 47.428648856401
		3 41.7267483589583
		4 31.8778123613383
		5 25.6593493527414
		6 20.31754411805
		7 18.1547613027856
		8 15.7287748381563
	};
	\addlegendentry{\Large $\text{L}^{\text{MSTM}} $}
	\addplot [semithick, red, dashed]
	table {%
		0 49.0961233864269
		1 48.7353802374612
		2 45.5892684769488
		3 39.7814737835273
		4 33.2137585869023
		5 24.4675295843483
		6 21.3394763822324
		7 19.1445935535033
		8 13.973620019562
	};
	\addlegendentry{\Large $\text{L}^{\text{SSTM}} $}
	\addplot [semithick, blue]
	table {%
		0 55.4115953353942
		1 53.8628141454491
		2 52.4575695353653
		3 48.980713470532
		4 43.2204931616751
		5 38.5069914460731
		6 35.4208624682231
		7 34.3450749768566
		8 31.9534891077711
	};
	\addlegendentry{\Large $\text{RL}^{\text{MSTM}} $}
	\addplot [semithick, red]
	table {%
		0 55.4115953353942
		1 53.2657746372965
		2 50.7167432537052
		3 44.5977392487042
		4 40.9496949268843
		5 33.9680119989935
		6 30.0628130013616
		7 32.02241631968
		8 26.0399276852317
	};
	\addlegendentry{\Large $\text{RL}^{\text{SSTM}} $}
	
	\nextgroupplot[
	tick align=outside,
	tick pos=left,
	title={\LARGE Random Sampling},
	x grid style={darkgray176},
	xlabel near ticks,
	xlabel={\LARGE Fraction of original lidar point cloud},
	xmin=0, xmax=8,
	xtick style={color=black},
	xtick={0,1,2,3,4,5,6,7,8},
	xticklabels={
		\Large 1,
		\Large \(\displaystyle \dfrac{1}{2}\),
		\Large \(\displaystyle \dfrac{1}{4}\),
		\Large \(\displaystyle \dfrac{1}{8}\),
		\Large \(\displaystyle \dfrac{1}{16}\),
		\Large \(\displaystyle \dfrac{1}{32}\),
		\Large \(\displaystyle \dfrac{1}{64}\),
		\Large \(\displaystyle \dfrac{1}{128}\),
		\Large \(\displaystyle \dfrac{1}{256}\)
	},
	y grid style={darkgray176},
	ymin=0, ymax=60,
	ytick style={color=black},
	yticklabels={
	},
	]
	\addplot [semithick, blue, dashed]
	table {%
		0 49.0152341518965
		1 44.5771763188908
		2 33.2730132117119
		3 29.3202511422097
		4 21.5882376187349
		5 11.9168258150581
		6 12.9506105079289
		7 7.61924751067394
		8 9.09090909090909
	};
	\addplot [semithick, red, dashed]
	table {%
		0 49.0152341518965
		1 38.9260483920233
		2 36.9627046154857
		3 27.2383192515014
		4 20.4222267789977
		5 13.862702950432
		6 9.16129940520188
		7 8.76080771249415
		8 9.27865718660698
	};
	\addplot [semithick, blue]
	table {%
		0 55.4115953353942
		1 53.1093168649834
		2 50.7138818210022
		3 45.8808316464705
		4 43.0996500655043
		5 41.5150497535399
		6 39.1565700403941
		7 31.9544555914788
		8 31.7345079824083
	};
	\addplot [semithick, red]
	table {%
		0 55.4115953353942
		1 50.8902748902342
		2 50.2832179707
		3 45.9811323077411
		4 41.5793861243018
		5 36.8473509515172
		6 34.3311580664081
		7 33.9391644237016
		8 28.3159401358828
	};
	
	\nextgroupplot[
	tick align=outside,
	tick pos=left,
	title={\LARGE Voxel-based Sampling},
	x grid style={darkgray176},
	xmin=0, xmax=8,
	xtick style={color=black},
	xtick={0,1,2,3,4,5,6,7,8},
	xticklabels={
		\Large 1,
		\Large \(\displaystyle \dfrac{1}{2}\),
		\Large \(\displaystyle \dfrac{1}{4}\),
		\Large \(\displaystyle \dfrac{1}{8}\),
		\Large \(\displaystyle \dfrac{1}{16}\),
		\Large \(\displaystyle \dfrac{1}{32}\),
		\Large \(\displaystyle \dfrac{1}{64}\),
		\Large \(\displaystyle \dfrac{1}{128}\),
		\Large \(\displaystyle \dfrac{1}{256}\)
	},
	y grid style={darkgray176},
	ymin=0, ymax=60,
	ytick style={color=black},
	yticklabels={},
	]
	\addplot [semithick, blue, dashed]
	table {%
		0 49.0152341518965
		1 39.1311201191968
		2 38.1556646523994
		3 34.7898643889949
		4 34.1674499324709
		5 25.538152788466
		6 22.2817667623503
		7 14.8899669071067
		8 12.436646143097
	};
	\addplot [semithick, red, dashed]
	table {%
		0 49.0152341518965
		1 41.9279531520783
		2 40.5459672621394
		3 36.4826657438761
		4 36.4826657438761
		5 28.4523479949822
		6 20.7058280572939
		7 15.869609306491
		8 10.1427253911505
	};
	\addplot [semithick, blue]
	table {%
		0 55.4115953353942
		1 51.5240803419798
		2 51.620142506343
		3 50.1439078280698
		4 49.9258863638712
		5 48.4438068356283
		6 45.0005802984301
		7 40.7370833260943
		8 39.9606863189168
	};
	\addplot [semithick, red]
	table {%
		0 55.4115953353942
		1 50.5009873250004
		2 50.2689210222329
		3 49.5508661535463
		4 49.5508661535463
		5 47.0400379567974
		6 40.4438738386649
		7 36.3437913209506
		8 36.7258257948929
	};
	\end{groupplot}
	
	\end{tikzpicture}
	\caption{Comparison between SSTM and MSTM for different lidar sampling strategies.}
	\label{fig:MultistageLidarRadar}
\end{figure*}

All thin-out strategies result in an increasing degradation of the detection performance. This is most pronounced when considering a random thin-out, which drops approximately linearly with each thin-out step. K-nearest neighbor and voxel-based sampling consistently perform better than random sampling due to keeping a higher point density in areas around objects, which supports object detection. For k-nearest neighbor sampling, the performance only drops by 0.6 percentage points between $\text{L}^{\text{SSTM}}_{\text{1 / knn}} $ and $\text{L}^{\text{SSTM}}_{\text{\sfrac{1}{2} / knn}} $ due to mostly ground points being removed in the first thin-out stage. For voxel-based sampling, after a sharp performance drop in the first thin-out stage, only a relatively slight performance drop is observable until the $\sfrac{1}{16}$ thin-out stage. In the first stages, the performance drop can mainly be attributed to the pedestrian class, which drops by 12.8 percentage points in the first thin-out stage.
In comparison, the detection performance of the car and cyclist classes only decreased by 0.7 and 3 percentage points, respectively. This can be explained by the voxel size of \SI{1}{\meter} in each dimension utilized in the voxel-based sampling. A single pedestrian only occupies a small number of these voxels. When the point cloud gets thinned out, the entire object's structure gets lost, making detection difficult.
In contrast, the object's structure and physical appearance can be adequately represented when it occupies more voxels, as observed in the car and cyclist classes. In the final thin-out stage of $\text{L}^{\text{SSTM}}$ and $\text{L}^{\text{MSTM}}$,  voxel-based sampling performs worse than k-nearest neighbor sampling because many voxels only consist of ground points or points of surrounding background objects.

The multi-stage training does not result in a useful knowledge transfer and, therefore, significant performance benefit for all considered thin-out strategies. 

A contrary behavior is observed on the combined radar + lidar point cloud. The MSTM consistently outperforms the SSTM when using k-nearest neighbor or voxel-based sampling. Knowledge from the dense point cloud can be transferred to the thin radar + lidar point cloud. Additionally, performance is consistently higher than training just on lidar, especially for voxel-based sub-sampling, which performs best at lower thin-out stages. At small thin-out stages, the voxel-based sub-sampling can still represent the whole object space and give meaningful environmental information. At the same time, the radar point cloud is sufficiently dense for object detection.

\subsection{Evaluation of MSTM with Last Radar-Only Step} \label{sec:MultiStageRadar}
This chapter analyses the performance of the MSTM when applied to radar data, as described in Section \ref{sec:MS_Methods}. The MSTM is evaluated for two different procedures. Utilizing only the lidar point cloud and utilizing the mixed radar + lidar point cloud in the first stages. The results of MSTM are shown in Table \ref{tab:MSRadar}. The lidar thin-out stages up to $\sfrac{1}{16}$ of the original lidar point cloud are considered due to the performance of $\text{L}^{\text{MSTM}}_{\text{1-\sfrac{1}{32} / rand}} $ dropping below $\text{R}^{\text{SSTM}} $. 

When considering the pre-training using just lidar data, it is observable that the overall performance drops in the SR bin. In contrast, the MR bin is improved for all thin-out methods. The best performance is achieved by utilizing the random thin-out strategy. It performs especially well for pedestrians and cyclists in the mid-range, due to the comparably small size of pedestrians and cyclists resulting in a worse radar reflection characteristic. In contrast these objects are detected well by the lidar sensor. Knowledge about the representation of objects can be transferred from the lidar point cloud to the radar point cloud. K-nearest neighbor and voxel-based thin-out do not improve the performance in the SR bin for the pedestrian and cyclist class but perform better in the short and mid-range for the car class which is explained by the large size of cars, resulting in a better representation of cars in the thin point cloud.

\begin{table*}[t]
	\centering
	\caption{3D object detection results on the VoD radar dataset trained using MSTM. All MSTM methods are trained on radar-only data in the last step. The best and second best results are marked in \textbf{bold} and \underline{underlined}, respectively.}
	\begin{tabular}{@{\extracolsep{1pt}}p{3cm}p{0.9cm}p{0.9cm}| p{0.9cm}p{0.9cm}p{0.9cm}p{0.9cm}p{0.9cm}p{0.9cm}}
		\toprule   
		{} & \multicolumn{2}{c|}{mAP} & \multicolumn{2}{c}{Car} & \multicolumn{2}{c}{Pedestrian} & \multicolumn{2}{c}{Cyclist} \\
		\cmidrule{2-3}
		\cmidrule{4-5}
		\cmidrule{6-7}
		\cmidrule{8-9}
		Training method & SR & MR & SR & MR & SR & MR & SR & MR \\ 
		\midrule
		$\text{R}^{\text{SSTM}}$ (baseline) & 36.7 & 11.9 & 45.2 & 18.1 & 17.1 & 7.4 & 47.7 & 10.2 \\
		\midrule
		$\text{L}^{\text{MSTM}}_{\text{1-\sfrac{1}{16}/ rand} }\rightarrow\text{R} $ & 36.4 & \textbf{16.1} & 44.9 & 20.0 & 17.5 & \textbf{9.8} & 46.8 & \textbf{18.9} \\
		$\text{L}^{\text{MSTM}}_{\text{1-\sfrac{1}{16} / knn} }\rightarrow\text{R} $ & 35.7 & 14.0 & 45.6 & \underline{23.2} & 16.7 & 7.1 & 44.8 & 11.7 \\
		$\text{L}^{\text{MSTM}}_{\text{1-\sfrac{1}{16}/ vox} }\rightarrow\text{R}    $ & 34.1 & 14.3 & \textbf{47.0} & 23.0 & 15.3 & 9.1 & 40.0 & 10.7 \\
		\midrule
		$\text{RL}^{\text{MSTM}}_{\text{1-\sfrac{1}{16}/ rand} }\rightarrow\text{R} $ & 35.6 & 14.9 & 44.1 & 20.2 & 17.8 & 7.2 & 44.9 & \underline{17.3} \\
		$\text{RL}^{\text{MSTM}}_{\text{1-\sfrac{1}{16} / knn} }\rightarrow\text{R} $ & \underline{38.2} & 14.7 & 45.5 & \textbf{23.9} & \textbf{18.8} & 8.2 & \underline{50.3} & 11.9 \\
		$\text{RL}^{\text{MSTM}}_{\text{1-\sfrac{1}{16} / vox} }\rightarrow\text{R} $ & \textbf{39.7} & \underline{15.4} & \underline{45.9} & 22.5 & \underline{18.4} & \underline{9.7} & \textbf{54.7} & 13.9 \\
		\bottomrule
	\end{tabular}
	\label{tab:MSRadar}
\end{table*}

When considering the pre-training with the mixed radar + lidar point cloud, a contrary observation is made compared to only using the lidar in pre-training. The overall performance for the random thinned-out lidar point cloud is worse with the mixed point cloud than when only considering the lidar point cloud. One consistent aspect is the good performance of the cyclists in the mid-range, which still surpasses all other methods, excluding $\text{L}^{\text{MSTM}}_{\text{1-\sfrac{1}{16} / ra}} $. However, the k-nearest neighbor and voxel-based thin-out strategies perform better with the mixed point cloud. The voxel-based thin-out strategy achieves the best performance. It performs best on objects in SR, mainly due to its outstanding performance in detecting cyclists, but is limited in the mid-range, getting surpassed by the MSTM with random thin-out. 
\textit{The MSTM with voxel-based thin-out can increase the object detection performance on the radar point cloud by 3 percentage points in the SR and 3.5 percentage points in the MR.}

\subsection{Evaluation of Cross-Modality KD} \label{sec:KD}
The configuration of the teacher's training data is of particular interest, as the teacher's performance directly influences the student's. The simplest choice is to train the teacher solely on the lidar point cloud. Section \ref*{sec:MultiStageLidar} and Section \ref*{sec:MultiStageRadar} show that mixing the radar and lidar point clouds can benefit radar-only object detection. Therefore, teachers trained on a mixed point clouds, in addition to the ones solely trained on lidar, are compared. Mixed point clouds containing $\sfrac{1}{4}$ of the original lidar point cloud are considered, as a closer representation of the teachers to the student's data can lead to a better performance. Thin out of $\sfrac{1}{4}$ is chosen as the radar + lidar detection performance with radar and $\sfrac{1}{8}$ of the lidar points is worse than the SSTM on only lidar data. Table \ref{tab:PretrainingKD} shows the performance of all utilized teachers.

Table \ref*{tab:PretrainingKD} shows the results of the teacher network for different training data configurations. It can be observed that, as expected, the mixed radar and lidar point cloud performs the best, only being surpassed by $\text{RL}^{\text{SSTM}}_{\text{\sfrac{1}{4} / knn}}$ on the cyclist class in short range. No model can be considered the overall second best among the other training sets. The performance varies between vehicle classes.

\begin{table*}[t]
	\centering
	\caption{3D object detection results of the teacher with different training sets. Each teacher is evaluated on the test set of the same data configuration as the training set. The best and second best results are marked in \textbf{bold} and \underline{underlined}, respectively.}
	\begin{tabular}{@{\extracolsep{1pt}}p{2.25cm}p{0.9cm}p{0.9cm}| p{0.9cm}p{0.9cm}p{0.9cm}p{0.9cm}p{0.9cm}p{0.9cm}}
		\toprule   
		{} & \multicolumn{2}{c|}{mAP} & \multicolumn{2}{c}{Car} & \multicolumn{2}{c}{Pedestrian} & \multicolumn{2}{c}{Cyclist} \\
		\cmidrule{2-3}
		\cmidrule{4-5}
		\cmidrule{6-7}
		\cmidrule{8-9}
		Data & SR & MR & SR & MR & SR & MR & SR & MR \\ 
		\midrule
		$\text{L}^{\text{SSTM}}_{\text{1 }}           $ & 56.5 & 30.0 & \underline{60.7} & \underline{42.4} & 40.1 & 19.5 & 68.5 & 28.1 \\
		$\text{RL}^{\text{SSTM}}_{\text{1 }}          $ & \textbf{61.6} & \textbf{45.1} & \textbf{61.9} & \textbf{46.5} & \textbf{44.4} & \textbf{34.0} & \underline{78.5} & \textbf{54.8}\\
		$\text{RL}^{\text{SSTM}}_{\text{\sfrac{1}{4} / rand}}   $ & 56.9 & 31.1 & 57.2 & 32.4 & 37.2 & 21.3 & 76.4 & 39.5 \\
		$\text{RL}^{\text{SSTM}}_{\text{\sfrac{1}{4} / knn}} $ & \underline{58.9} & 36.9 & 54.4 & 31.9 & \underline{43.0} & \underline{30.9} & \textbf{79.5} & \underline{47.9} \\
		$\text{RL}^{\text{SSTM}}_{\text{\sfrac{1}{4} / vox}}    $ & 56.1 & \underline{39.3} & 60.1 & 41.6 & 32.8 & 28.8 & 75.5 & 47.5 \\
		\bottomrule
	\end{tabular}
	\label{tab:PretrainingKD}
\end{table*}

The KD is evaluated individually for each KD method and teacher training set. A joint KD is also considered comprised of all three KD losses. 
All student networks are initialized (Init) with the weights of the teacher network, as it has been shown to improve the student's performance \cite{Yang2022KDOpenPCDet}. Additionally, for the transfer learning between datasets, the pre-training utilizing MSTM in Section \ref*{sec:MultiStageRadar} has been shown to improve the performance on the radar dataset. 

The results utilizing the KD are shown in Table \ref{tab:KnowledgeDistillation}. Just the initialization of the student network with the teacher's weights already results in a performance benefit in the SR with $ \text{RL}^{\text{SSTM}}_{\text{1}} $ and  $\text{RL}^{\text{SSTM}}_{\text{\sfrac{1}{4} / rand}} $ as teachers and in the MR with all teachers. This is overall only surpassed by the $\text{RL}^{\text{MSTM}}_{\text{1-\sfrac{1}{16} / vox}}\rightarrow\text{R} $ showing that the MSTM is substantially better than a simple initialization. 

\begin{table*}[t]
	\centering
	\caption{3D object detection results on the VoD radar dataset utilizing the different KDs and teachers trained on different data configurations. Only the mAP over all classes is specified. For each distillation method and range bin, the best and second best results are marked in \textbf{bold} and \underline{underlined}, respectively. The best result for each teacher are marked in \textcolor{cyan}{cyan} for the SR bin and \textcolor{magenta}{magenta} for the MR bin.}
	\begin{tabular}{@{\extracolsep{1pt}}p{2.25cm}p{0.85cm}p{0.85cm}| p{0.85cm}p{0.85cm}|p{0.85cm}p{0.85cm}|p{0.85cm}p{0.85cm}|p{0.85cm}p{0.85cm}}
		\toprule   
		{} & \multicolumn{2}{c|}{Init only} & \multicolumn{2}{c}{Logit-KD} & \multicolumn{2}{c}{Feature-KD} & \multicolumn{2}{c}{Label-KD} & \multicolumn{2}{c}{Joint-KD} \\ 
		{} & \multicolumn{2}{c|}{($\rightarrow\text{R}^{\text{}}$)} & \multicolumn{2}{c}{($\rightarrow\text{R}^{\text{log}}$)} & \multicolumn{2}{c}{($\rightarrow\text{R}^{\text{feat}}$)} & \multicolumn{2}{c}{($\rightarrow\text{R}^{\text{lab}}$)} & \multicolumn{2}{c}{($\rightarrow\text{R}^{\text{Joint}}$)} \\
		\cmidrule{2-3}
		\cmidrule{4-5}
		\cmidrule{6-7}
		\cmidrule{8-9}
		\cmidrule{10-11}
		Teacher data & SR & MR & SR & MR & SR & MR & SR & MR & SR & MR \\ 
		\midrule
		$\text{R}^{\text{SSTM}}_{\text{1}}$ & 36.7 & 11.9 & \;- & \;- & \;- & \;- & \;- & \;- & \;- & \;- \\
		\midrule
		$\text{L}^{\text{SSTM}}_{\text{1 }}           $ & 34.9 & 13.5 & 32.8 & 12.6 & 34.8 & 13.6 & \textcolor{cyan}{\underline{36.6}} & \textcolor{magenta}{\underline{13.7}} & 36.0 &  \textbf{13.5} \\
		$\text{RL}^{\text{SSTM}}_{\text{1 }}          $ & \textbf{39.0} & \textcolor{magenta}{\underline{14.8}} & \textcolor{cyan}{\textbf{39.2}} & \underline{13.5} & \underline{39.1} & \underline{13.8} & \textbf{38.8} & \underline{13.7} & \textbf{38.9} & 12.7  \\
		$\text{RL}^{\text{SSTM}}_{\text{\sfrac{1}{4} / rand}}     $ & \textbf{39.0} & \textcolor{magenta}{14.2} & \underline{38.9} & 11.6 & \textcolor{cyan}{\textbf{39.4}} & 11.8 & 34.0 & 12.2 & \underline{36.4} &  12.3 \\
		$\text{RL}^{\text{SSTM}}_{\text{\sfrac{1}{4} / knn}}      $ & 34.8 & \textcolor{magenta}{14.6} & 37.2 & \textbf{14.7} & \textcolor{cyan}{36.6} & 13.0 & 33.8 & 10.9 & 32.9 & \underline{12.9}  \\
		$\text{RL}^{\text{SSTM}}_{\text{\sfrac{1}{4} / vox}}      $ & \underline{35.2} & \textbf{15.3} & 35.6 & 13.3 & \textcolor{cyan}{37.1} & \textcolor{magenta}{\textbf{15.8}} & 35.3 & \textbf{14.3} & 32.9 & 9.7  \\
		\bottomrule
	\end{tabular}
	\label{tab:KnowledgeDistillation}
\end{table*}

For label-KD, the best-performing models are the ones where the teacher performs the best. Specifically $\text{L}^{\text{SSTM}}_{\text{1}} \rightarrow\text{R}^{\text{lab}} $, which is only surpassed by \break $\text{RL}^{\text{SSTM}}_{\text{\sfrac{1}{4} / knn}} \rightarrow\text{R}^{\text{lab}}$ in the MR. For worse-performing teachers, using label-KD loss does not result in a performance benefit due to it replacing the ground truth label loss. Feature-KD requires a teacher who learns features from radar data. This is observable in the SR performance of models, where the teacher is trained on mixed radar + lidar data. Logit-KD works well on teacher datasets which closely resemble the radar point cloud and perform well on the teacher set. This results in good performance of $ \text{RL}^{\text{SSTM}}_{\text{1}}\rightarrow\text{R}^{\text{log}} $. Besides the KD method the thin-out method utilized in the teacher's training influences the student's performance. Random sampling results in good performance in the SR, while voxel-based sampling results in good MR performance. This is somewhat contrary to what has been observed in the MSTM, where random sampling performs the best in the MR and can be explained by the different thin-out stages utilized in the MSTM and KD. Further qualitative results, as well as detailed quantitative results are given in the supplementary. \textit{Overall, initializing the student with the teacher's parameters yields good performance, with further enhancements primarily achievable through feature-KD with a teacher trained on mixed radar + lidar data.}

\subsection{Evaluation on DSVT as a Transformer-Based Object Detector}
Table \ref{tab:VoxelRCNN} shows the results of the MSTM and KD on DSVT-P \cite{wang2023dsvt} for selected methods. Similar effects, as observed for PointPillars, are seen on DSVT-P. Initializing the student with the teacher's weights and the MSTM led to a performance benefit. However, contrary to PointPillars, the KD approach does not contribute to any improvements for DSVT-P.

\begin{table*}[t]
	\centering
	\caption{Radar-only detection performance utilizing DSVT-P as an object detector. The best and second best results are marked in \textbf{bold} and \underline{underlined}, respectively.}
	\begin{tabular}{@{\extracolsep{1pt}}p{3cm}p{0.9cm}p{0.9cm}| p{0.9cm}p{0.9cm}p{0.9cm}p{0.9cm}p{0.9cm}p{0.9cm}}
		\toprule   
		{} & \multicolumn{2}{c|}{mAP} & \multicolumn{2}{c}{Car} & \multicolumn{2}{c}{Pedestrian} & \multicolumn{2}{c}{Cyclist} \\
		\cmidrule{2-3}
		\cmidrule{4-5}
		\cmidrule{6-7}
		\cmidrule{8-9}
		Training method & SR & MR & SR & MR & SR & MR & SR & MR \\ 
		\midrule
		$\text{R}^{\text{SSTM}}                                                           $ & 38.3 & 13.0 & 42.5 & \underline{20.2} & 21.9 & \textbf{12.3} & \underline{50.5} & 6.6 \\
		\midrule
    $\text{RL}^{\text{MSTM}}_{\text{1-\sfrac{1}{16} / vox}}\rightarrow\text{R}            $ & \textbf{41.5} & \textbf{15.6} & \textbf{47.8} & \textbf{23.6} & \textbf{24.5} & 11.0 & \textbf{52.3} & \textbf{12.4} \\
		\midrule
    $\text{RL}^{\text{SSTM}}_{\text{\sfrac{1}{4} / vox}}\rightarrow\text{R}               $ & \underline{38.8} & \underline{13.2} & \underline{44.9} & 18.8 & \underline{23.3} & \underline{11.2} & 48.2 & 9.7 \\
    $\text{RL}^{\text{SSTM}}_{\text{\sfrac{1}{4} / vox}}\rightarrow\text{R}^{\text{feat}} $ & 34.7 & 11.3 & 44.0 & 8.8 & 20.5 & 2.8 & 39.7 & \underline{12.7} \\
		\bottomrule
	\end{tabular}
	\label{tab:VoxelRCNN}
\end{table*}

\subsection{Limitations}
The best-performing methods in this work apply only to detectors that share an architecture with the target model, as a direct transfer of weights is performed for the best performance. Different models may require different levels and steps in the thinning process. These choices are additional tuning parameters that must be selected appropriately to maximize the benefit of transferring knowledge. 

\section{Conclusion}
\label{sec:conclusion}
In this paper, we investigated two methods to transfer knowledge from lidar-based object detectorss to radar-only object detection. First, MSTM with sequential sub-sampling of the lidar point cloud, and second, a KD-based approach. For the MSTM, we have investigated three thin-out strategies for the lidar point cloud. These thin-out strategies are also analyzed for the training of the KD teacher network. Both methods can substantially improve the object detection performance of a radar-only object detector. The MSTM with voxel-based thin-out performs the best overall and can improve detection performance by up to 3.5 percentage points. For the KD methods, it is shown that initializing the student with the teacher's parameters, especially a teacher trained on mixed lidar and radar data, can improve the object detection performance on radar-only data, with further enhancement primarily achieved by utilizing feature-KD.

In \textit{future work}, the applicability to further 3D object detection networks and the behavior with more advanced knowledge distillation like the ones utilized by \cite{Klingner2023X3KD} could be investigated. Due to different effects being noticed with the MSTM and the KD methods, combining both methods could be investigated by utilizing the MSTM as a teacher. To overcome the limitations of choosing a strict thin-out strategy, a learnable point cloud thin-out method \cite{lang2020samplenet, Zhang_2022_CVPR} can be used.

%
%

\title{Supplementary Material for: \\ LEROjD:  \underline{L}idar \underline{E}xtended \underline{R}adar-Only \underline{O}b\underline{j}ect \underline{D}etection} 

\titlerunning{Supplementary material for LEROjD}

\author{Patrick Palmer\inst{1} \and
	Martin Krüger\inst{1} \and
	Stefan Schütte\inst{1} \and
	Richard Altendorfer\inst{2} \and 
	Ganesh Adam\inst{2} \and 
	Torsten Bertram\inst{1}}

\authorrunning{P.~Palmer et al.}

\institute{Institute of Control Theory and Systems Engineering, TU Dortmund University
	\\
	\and
	ZF Group
	}
\maketitle

\appendix
\section{Detailed Experimental Results}
Table \ref{tab:CMKD} extends Table 4 in the original paper by evaluating the detection performance per vehicle class. General trends shown in the main paper on the mAP are also observable across the three considered vehicle classes. Notable are:

\begin{itemize}
	\item The best performance of $\text{RL}^{\text{SSTM}}_{\text{\sfrac{1}{4} / vox}}\rightarrow\text{R}$ in the cyclist class in the MR.
	\item The consistently good performance of teacher $\text{RL}^{\text{SSTM}}_{1}$ across different knowledge distillation methods for the pedestrian and cyclist class.
	\item The best performance of $\text{RL}^{\text{SSTM}}_{\text{1}}\rightarrow\text{R}$ in the car class in the SR.
	\item The best performance of $\text{RL}^{\text{SSTM}}_{\text{1}}\rightarrow\text{R} $ and $\text{L}^{\text{SSTM}}_{1}\rightarrow\text{R}_{\text{feat}} $ for the car class in the MR.
\end{itemize}

Tables \ref{tab:MSTM1} to \ref{tab:MSTM3} show the object detection results for lidar-only SSTM shown in Fig. 3 in the main paper, split into the two considered range areas and vehicle classes. While the performance for most training methods decreases monotonically with each thin-out step, there are a few exceptions. For example, while the performance of $\text{L}^{\text{SSTM}}_{\text{\sfrac{1}{4} / knn}}$ is worse than $\text{L}^{\text{SSTM}}_{\text{\sfrac{1}{2} / knn}}$ for cars and pedestrians in the SR, it is better for cyclists by 3.9 percentage points. One reason for this is the variance between runs in training due to random initialization of the model, which has been partly mitigated by using three random initializations of each model. Another reason is the performance tradeoff between classes. Training the model for optimal performance over all classes can lead to increased performance for one class at the cost of performance in the other. Comparing the performance between vehicle classes and the thin-out method, it is observable that the voxel-based sampling suffers in the pedestrian class due to the comparably small size of pedestrians. For the different range areas, it can be observed that random sampling is not well suited for the MR due to the point cloud losing its structure in this area when randomly thinned-out. 

\section{Qualitative Evaluation of MSTM and Cross-Modal Knowledge Distillation}

\begin{figure}
	\centering
	\begin{subfigure}{0.75\textwidth}
			\includegraphics[width=\textwidth]{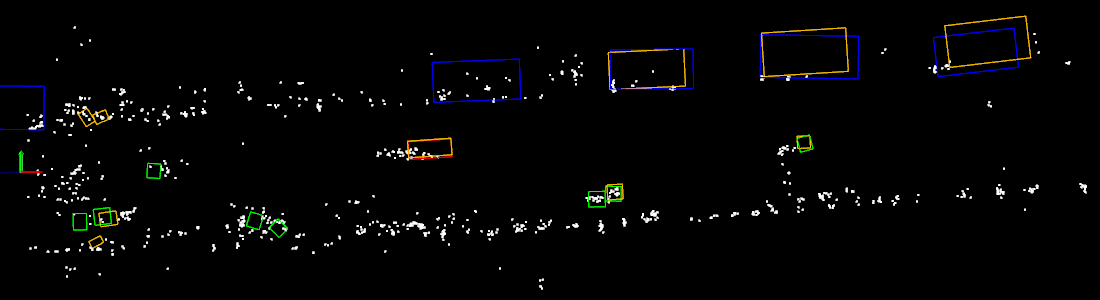}
			\caption{$\text{R}^{\text{SSTM}}_1 $}
			\label{fig:first}
	\end{subfigure}
	\hfill
	\begin{subfigure}{0.75\textwidth}
			\includegraphics[width=\textwidth]{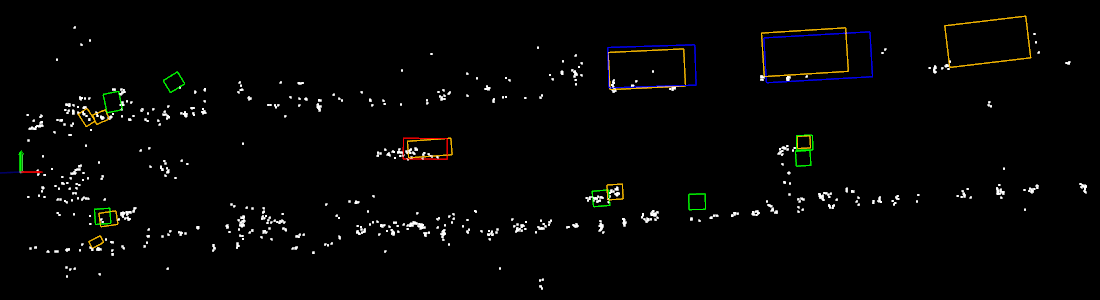}
			\caption{$\text{RL}^{\text{MSTM}}_{\text{1-\sfrac{1}{16} / vox} }\rightarrow\text{R}$}
			\label{fig:second}
	\end{subfigure}
	\hfill
	\begin{subfigure}{0.75\textwidth}
			\includegraphics[width=\textwidth]{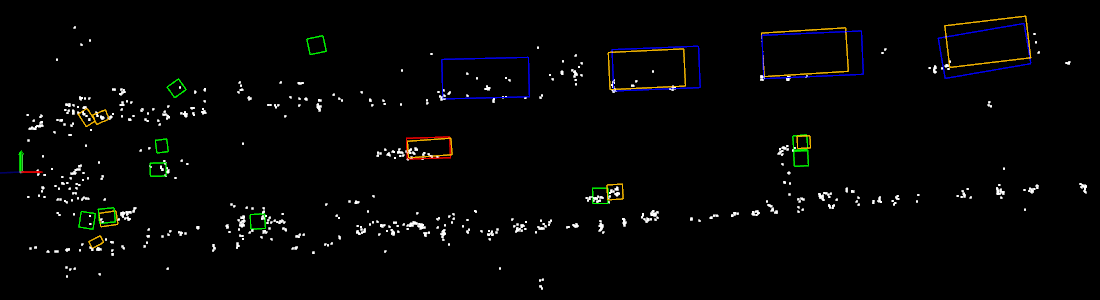}
			\caption{$\text{RL}^{\text{SSTM}}_{\text{1}}\rightarrow\text{R}$}
			\label{fig:third}
	\end{subfigure}
	\begin{subfigure}{0.75\textwidth}
		\includegraphics[width=\textwidth]{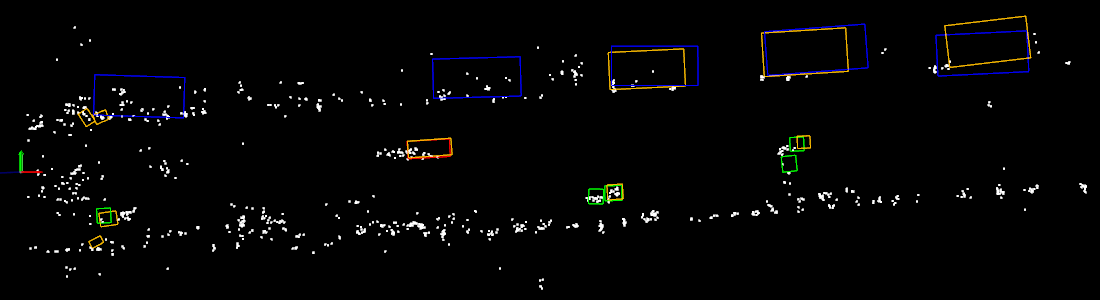}
		\caption{$\text{RL}^{\text{SSTM}}_{\text{1}}\rightarrow\text{R}_{\text{logit}}$}
		\label{fig:fourth}
\end{subfigure}
\hfill
\begin{subfigure}{0.75\textwidth}
		\includegraphics[width=\textwidth]{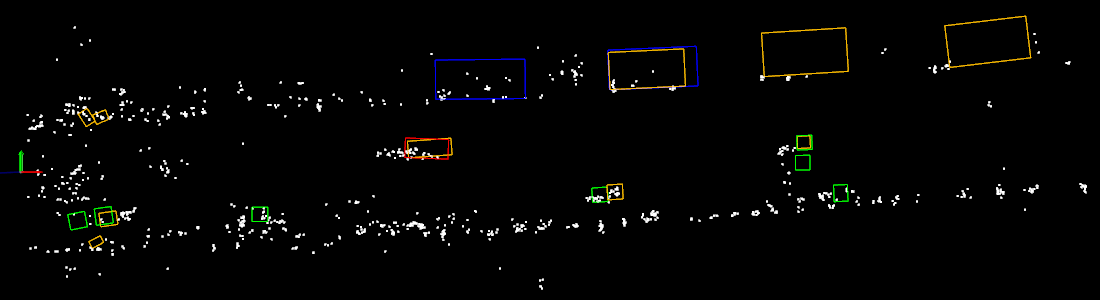}
		\caption{$\text{RL}^{\text{SSTM}}_{\text{\sfrac{1}{4} / vox}}\rightarrow\text{R}_{\text{}}$}
		\label{fig:firth}
\end{subfigure}
\hfill
\begin{subfigure}{0.75\textwidth}
		\includegraphics[width=\textwidth]{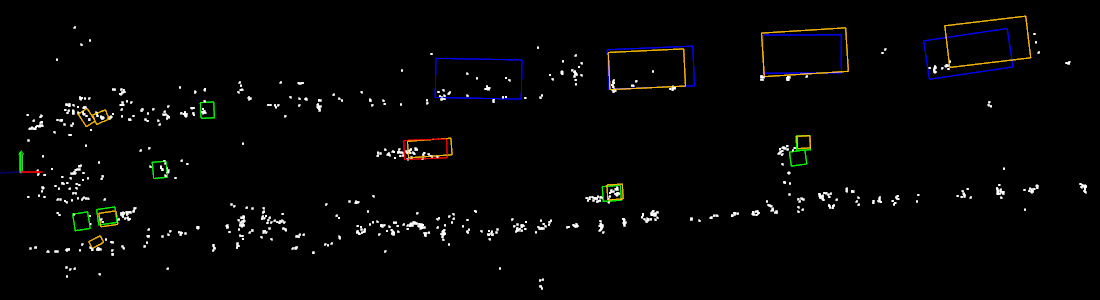}
		\caption{$\text{RL}^{\text{SSTM}}_{\text{\sfrac{1}{4} / vox}}\rightarrow\text{R}_{\text{feat}}$}
		\label{fig:sixt}
\end{subfigure}
					
	\caption{Detection results on radar-only data utilizing selected training methods in bird's-eye view representation. The white points are single 3D radar measurements, of a point cloud accumulated over 5 frames. Orange rectangles represent ground truth annotations for all object classes. Blue, red, and green rectangles visualize the detection of cars, cyclists, and pedestrians, respectively.}
	\label{fig:figures}
\end{figure}

\begin{figure}
		\centering
		\begin{subfigure}{0.75\textwidth}
				\includegraphics[width=\textwidth]{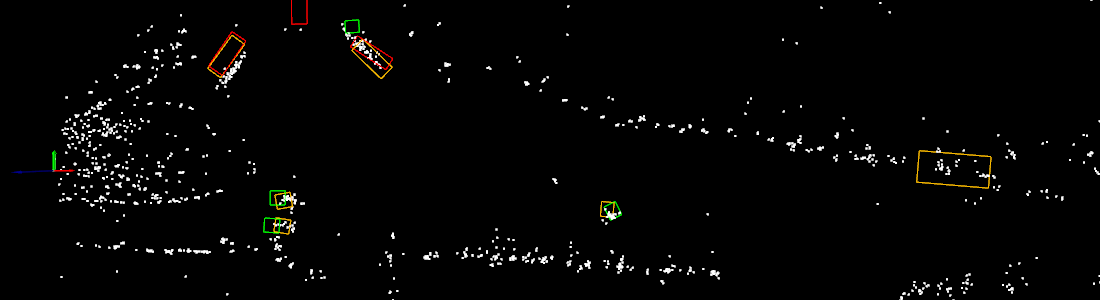}
				\caption{$\text{R}^{\text{SSTM}}_1 $}
				\label{fig:first1}
		\end{subfigure}
		\hfill
		\begin{subfigure}{0.75\textwidth}
				\includegraphics[width=\textwidth]{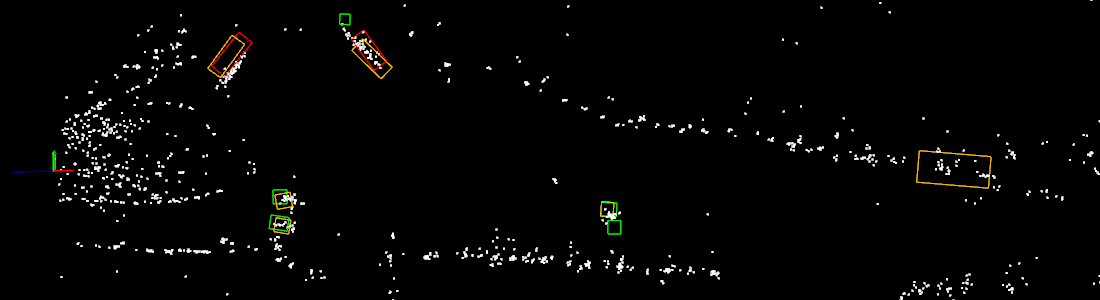}
				\caption{$\text{RL}^{\text{MSTM}}_{\text{1-\sfrac{1}{16} / vox} }\rightarrow\text{R}$}
				\label{fig:second2}
		\end{subfigure}
		\hfill
		\begin{subfigure}{0.75\textwidth}
				\includegraphics[width=\textwidth]{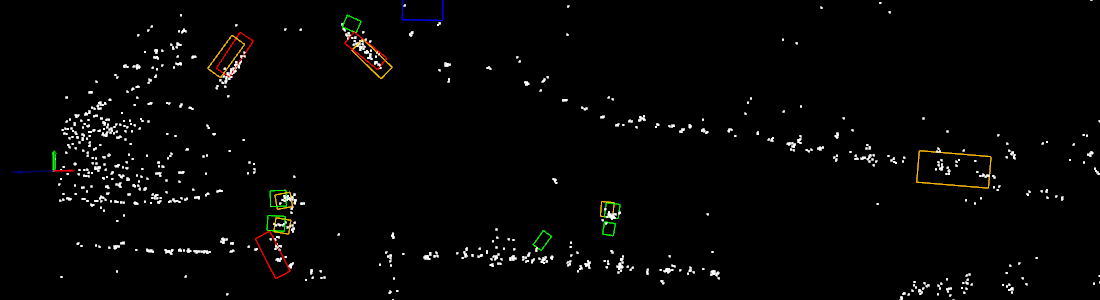}
				\caption{$\text{RL}^{\text{SSTM}}_{\text{1}}\rightarrow\text{R}$}
				\label{fig:third3}
		\end{subfigure}
		\begin{subfigure}{0.75\textwidth}
			\includegraphics[width=\textwidth]{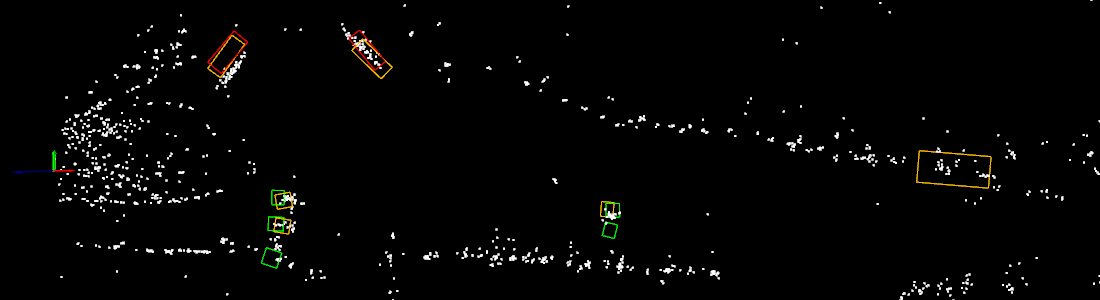}
			\caption{$\text{RL}^{\text{SSTM}}_{\text{1}}\rightarrow\text{R}_{\text{logit}}$}
			\label{fig:fourth4}
	\end{subfigure}
	\hfill
	\begin{subfigure}{0.75\textwidth}
			\includegraphics[width=\textwidth]{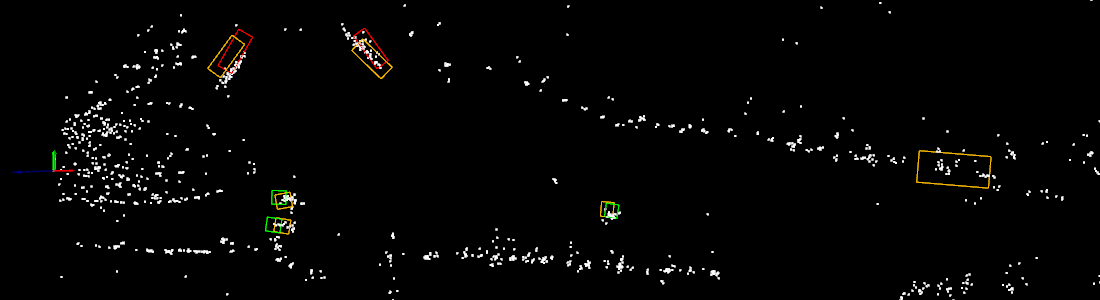}
			\caption{$\text{RL}^{\text{SSTM}}_{\text{\sfrac{1}{4} / vox}}\rightarrow\text{R}_{\text{}}$}
			\label{fig:firth5}
	\end{subfigure}
	\hfill
	\begin{subfigure}{0.75\textwidth}
			\includegraphics[width=\textwidth]{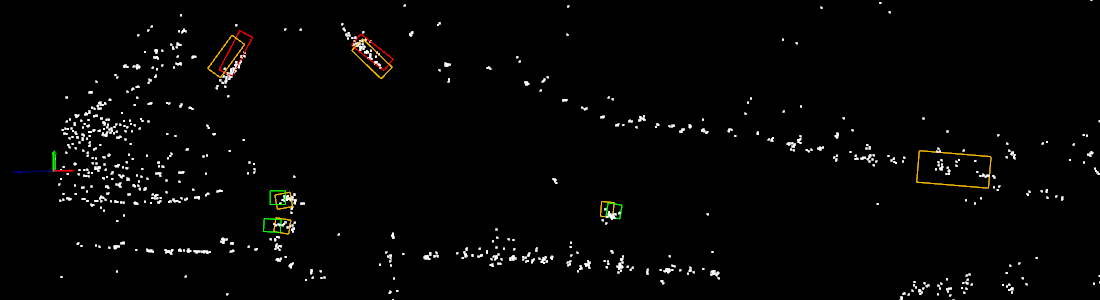}
			\caption{$\text{RL}^{\text{SSTM}}_{\text{\sfrac{1}{4} / vox}}\rightarrow\text{R}_{\text{feat}}$}
			\label{fig:sixt6}
	\end{subfigure}
						
		\caption{Detection results on radar-only data utilizing selected training methods in bird's-eye view representation. The white points are single 3D radar measurements, of a point cloud accumulated over 5 frames. Orange rectangles represent ground truth annotations for all object classes. Blue, red, and green rectangles visualize the detection of cars, cyclists, and pedestrians, respectively.}
		\label{fig:figures1}
\end{figure}

Fig. \ref{fig:figures} and \ref{fig:figures1} show annotated ground truth and detection outputs on radar-only data overlaid on a bird's-eye view representation of the radar-only point cloud for selected methods of MSTM and cross-modal knowledge distillation. For the baseline method $\text{R}^{\text{SSTM}}_1$, many false positive detections, especially for the pedestrian class, are observable. {$\text{RL}^{\text{MSTM}}_{\text{1-\sfrac{1}{16} / vox} }\rightarrow\text{R}$} reduced the number of false positives to 4 compared to the baselines 7 in Fig. \ref{fig:figures} while retaining most true positives. Regarding cross-modal knowledge distillation, it is observed that an initialization with $\text{RL}^{\text{SSTM}}_{\text{1}}$ results in a high number of false positives and a higher number of true positives. Utilizing logit-KD, the number of false positives can be reduced from 8 to 4 in Fig. \ref{fig:figures} and from 5 to 2 in Fig. \ref{fig:figures1}. $\text{RL}^{\text{SSTM}}_{\text{\sfrac{1}{4} / vox}}\rightarrow\text{R}$ produces a comparatively low number of false positives of 5 in Fig. \ref{fig:figures} and 0 in Fig. \ref{fig:figures1}, with minor gains in true positive detections utilizing feature-KD.

\begin{table*}
	\centering
	\caption{Object detection results for all cross-modal KD methods. The best and second best results are marked in bold and underlined, respectively.}
	\begin{tabular}{@{\extracolsep{1pt}}p{3.1cm}|p{0.9cm}p{0.9cm}| p{0.87cm}p{0.87cm}p{0.87cm}p{0.87cm}p{0.87cm}p{0.87cm}}
		\toprule   
		{} & \multicolumn{2}{c|}{mAP} & \multicolumn{2}{c}{Car} & \multicolumn{2}{c}{Pedestrian} & \multicolumn{2}{c}{Cyclist} \\
		\cmidrule{2-3}
		\cmidrule{4-5}
		\cmidrule{6-7}
		\cmidrule{8-9}
		Training method & SR & MR & SR & MR & SR & MR & SR & MR \\ 
		\midrule
		$\text{L}^{\text{SSTM}}_{\text{1}}\rightarrow\text{R}                                   $  & 34.9 & 13.6 & 43.9 & 20.1 & 16.9 & \hspace{0.15cm}8.5 & 43.9 & 12.1 \\
		$\text{L}^{\text{SSTM}}_{\text{1}}\rightarrow\text{R}^{\text{feat}}                     $  & 34.8 & 14.1 & 44.3 & \underline{21.7} & 15.3 & \hspace{0.15cm}7.0 & 44.8 & 13.7 \\
		$\text{L}^{\text{SSTM}}_{\text{1}}\rightarrow\text{R}^{\text{log}}                    $  & 32.8 & 12.6 & 42.9 & 18.2 & 16.0 & \hspace{0.15cm}8.7 & 39.6 & 10.9 \\
		$\text{L}^{\text{SSTM}}_{\text{1}}\rightarrow\text{R}^{\text{lab}}                    $  & 36.6 & 13.7 & 46.0 & 21.1 & 18.5 & \hspace{0.15cm}8.3 & 45.4 & 11.8 \\
		$\text{L}^{\text{SSTM}}_{\text{1}}\rightarrow\text{R}^{\text{joint}}                   $  & 36.0 & 13.5 & 46.7 & 20.6 & 16.9 & \hspace{0.15cm}8.7 & 44.4 & 11.3 \\
		\midrule
		$\text{RL}^{\text{SSTM}}_{\text{1}}\rightarrow\text{R}                                  $  & 39.0 & 14.8 & 44.5 & \textbf{21.8} & \underline{19.6} & \hspace{0.15cm}9.6 & 52.8 & 13.0 \\
		$\text{RL}^{\text{SSTM}}_{\text{1}}\rightarrow\text{R}^{\text{feat}}                    $  & 39.1 & 13.8 & 44.0 & 20.7 & \underline{19.6} & \hspace{0.15cm}8.0 & \underline{53.8} & 12.7 \\
		$\text{RL}^{\text{SSTM}}_{\text{1}}\rightarrow\text{R}^{\text{log}}                   $  & \underline{39.2} & 13.5 & 44.1 & 18.9 & \textbf{19.8} & \hspace{0.15cm}9.2 & \textbf{53.9} & 12.5 \\
		$\text{RL}^{\text{SSTM}}_{\text{1}}\rightarrow\text{R}^{\text{lab}}                   $  & 38.8 & 13.7 & 46.8 & 20.9 & 19.2 & \hspace{0.15cm}8.1 & 50.3 & 12.2 \\
		$\text{RL}^{\text{SSTM}}_{\text{1}}\rightarrow\text{R}^{\text{joint}}                  $  & 38.9 & 12.7 & 45.5 & 21.2 & 19.4 & \hspace{0.15cm}8.9 & 51.8 & \hspace{0.15cm}7.9 \\
		\midrule
		$\text{RL}^{\text{SSTM}}_{\text{\sfrac{1}{4} / rand}}\rightarrow\text{R}                $  & 39.0 & 14.2 & \underline{47.6} & 20.4 & 18.3 & \textbf{11.5} & 51.0 & 10.7 \\
		$\text{RL}^{\text{SSTM}}_{\text{\sfrac{1}{4} / rand}}\rightarrow\text{R}^{\text{feat}}  $  & \textbf{39.4} & 11.8 & \textbf{49.2} & 18.4 & 18.2 & \hspace{0.15cm}8.1 & 50.7 & \hspace{0.15cm}9.0 \\
		$\text{RL}^{\text{SSTM}}_{\text{\sfrac{1}{4} / rand}}\rightarrow\text{R}^{\text{log}} $  & 38.9 & 11.6 & 45.9 & 20.1 & 18.0 & \hspace{0.15cm}5.7 & 52.8 & \hspace{0.15cm}8.9 \\
		$\text{RL}^{\text{SSTM}}_{\text{\sfrac{1}{4} / rand}}\rightarrow\text{R}^{\text{lab}} $  & 34.0 & 12.2 & 43.8 & 19.2 & 15.2 & \hspace{0.15cm}5.0 & 42.9 & 12.3 \\
		$\text{RL}^{\text{SSTM}}_{\text{\sfrac{1}{4} / rand}}\rightarrow\text{R}^{\text{joint}}$  & 36.4 & 12.3 & 45.2 & 19.1 & 17.2 & \hspace{0.15cm}7.7 & 46.7 & 10.1 \\
		\midrule
		$\text{RL}^{\text{SSTM}}_{\text{\sfrac{1}{4} / knn}}\rightarrow\text{R}                 $  & 34.8 & 14.6 & 43.6 & 20.5 & 16.6 & \hspace{0.15cm}9.9 & 44.2 & 13.4 \\
		$\text{RL}^{\text{SSTM}}_{\text{\sfrac{1}{4} / knn}}\rightarrow\text{R}^{\text{feat}}   $  & 36.6 & 13.0 & 43.9 & 18.5 & 18.1 & \hspace{0.15cm}7.2 & 47.8 & 13.2 \\
		$\text{RL}^{\text{SSTM}}_{\text{\sfrac{1}{4} / knn}}\rightarrow\text{R}^{\text{log}}  $  & 37.2 & 14.7 & 44.1 & 21.2 & 17.8 & \hspace{0.15cm}7.9 & 49.8 & \underline{15.0} \\
		$\text{RL}^{\text{SSTM}}_{\text{\sfrac{1}{4} / knn}}\rightarrow\text{R}^{\text{lab}}  $  & 33.8 & 10.9 & 44.0 & 19.8 & 16.8 & \hspace{0.15cm}5.8 & 40.7 & \hspace{0.15cm}7.2 \\
		$\text{RL}^{\text{SSTM}}_{\text{\sfrac{1}{4} / knn}}\rightarrow\text{R}^{\text{joint}} $  & 32.9 & 12.9 & 42.9 & 17.8 & 14.8 & \hspace{0.15cm}8.6 & 41.0 & 12.2 \\
		\midrule
		$\text{RL}^{\text{SSTM}}_{\text{\sfrac{1}{4} / vox}}\rightarrow\text{R}                 $  & 35.2 & \underline{15.3} & 44.1 & 17.6 & 17.0 & \hspace{0.15cm}8.6 & 44.7 & \textbf{19.6} \\
		$\text{RL}^{\text{SSTM}}_{\text{\sfrac{1}{4} / vox}}\rightarrow\text{R}^{\text{feat}}   $  & 37.1 & \textbf{15.8} & 44.0 & 21.1 & 17.7 & \underline{11.3} & 49.8 & 14.7 \\
		$\text{RL}^{\text{SSTM}}_{\text{\sfrac{1}{4} / vox}}\rightarrow\text{R}^{\text{log}}  $  & 35.6 & 13.2 & 45.7 & 20.0 & 16.8 & \hspace{0.15cm}7.6 & 44.3 & 12.1 \\
		$\text{RL}^{\text{SSTM}}_{\text{\sfrac{1}{4} / vox}}\rightarrow\text{R}^{\text{lab}}  $  & 35.3 & 14.3 & 44.4 & 19.4 & 16.3 & 10.1 & 45.3 & 12.5 \\
		$\text{RL}^{\text{SSTM}}_{\text{\sfrac{1}{4} / vox}}\rightarrow\text{R}^{\text{joint}} $  & 32.8 & 9.7  & 46.2 & 15.5 & 14.0 & \hspace{0.15cm}6.4 & 38.4 & \hspace{0.15cm}7.0 \\
		\bottomrule
	\end{tabular}
	\label{tab:CMKD}
\end{table*}

\begin{table*}
	\centering
	\caption{Object detection results for all thin-out steps of $\text{L}^{\text{SSTM}}_{\text{knn}}$.}
	\begin{tabular}{@{\extracolsep{1pt}}p{2.8cm}p{0.9cm}p{0.9cm}| p{0.9cm}p{0.9cm}p{0.9cm}p{0.9cm}p{0.9cm}p{0.9cm}}
		\toprule   
		{} & \multicolumn{2}{c|}{mAP} & \multicolumn{2}{c}{Car} & \multicolumn{2}{c}{Pedestrian} & \multicolumn{2}{c}{Cyclist} \\
		\cmidrule{2-3}
		\cmidrule{4-5}
		\cmidrule{6-7}
		\cmidrule{8-9}
		Training method & SR & MR & SR & MR & SR & MR & SR & MR \\ 
		\midrule
		$\text{L}^{\text{SSTM}}_{\text{1} }                        $ & 56.3 & 34.0 & 59.8 & 46.3 & 42.2 & 17.5 & 66.7 & 38.3 \\
		\midrule
		$\text{L}^{\text{SSTM}}_{\text{\sfrac{1}{2} / knn} }       $ & 55.5 & 32.3 & 59.5 & 39.9 & 40.4 & 23.1 & 66.7 & 35.7 \\
		$\text{L}^{\text{SSTM}}_{\text{\sfrac{1}{4} / knn} }       $ & 53.9 & 31.9 & 54.4 & 40.0 & 36.7 & 23.7 & 70.5 & 32.1 \\
		$\text{L}^{\text{SSTM}}_{\text{\sfrac{1}{8} / knn} }       $ & 51.1 & 23.3 & 53.6 & 29.9 & 35.1 & 13.3 & 64.5 & 26.8 \\
		$\text{L}^{\text{SSTM}}_{\text{\sfrac{1}{16} / knn} }      $ & 42.9 & 13.0 & 50.0 & 19.7 & 28.4 & \hspace{0.15cm}7.0 & 50.2 & 12.2 \\
		$\text{L}^{\text{SSTM}}_{\text{\sfrac{1}{32} / knn} }      $ & 35.4 & 9.0 & 43.6 & 14.8 & 24.3 & \hspace{0.15cm}3.0 & 38.4 & \hspace{0.15cm}9.1 \\
		$\text{L}^{\text{SSTM}}_{\text{\sfrac{1}{64} / knn} }      $ & 27.7 & 3.8 & 32.9 & \hspace{0.15cm}9.1 & 20.6 & \hspace{0.15cm}2.3 & 29.7 & \hspace{0.15cm}0.1\\
		$\text{L}^{\text{SSTM}}_{\text{\sfrac{1}{128} / knn} }     $ & 26.2 & 4.9 & 39.0 & 10.2 & 15.0 & \hspace{0.15cm}4.5 & 24.5 & \hspace{0.15cm}0.0 \\
		$\text{L}^{\text{SSTM}}_{\text{\sfrac{1}{256} / knn} }     $ & 21.9 & 3.6 & 33.2 & \hspace{0.15cm}9.1 & 14.9 & \hspace{0.15cm}0.6 & 17.6 & \hspace{0.15cm}1.0 \\
		\bottomrule
	\end{tabular}
	\label{tab:MSTM1}
\end{table*}

\begin{table*}
	\centering
	\caption{Object detection results for all thin-out steps of $\text{L}^{\text{SSTM}}_{\text{rand}}$.}
	\begin{tabular}{@{\extracolsep{1pt}}p{2.8cm}p{0.9cm}p{0.9cm}| p{0.9cm}p{0.9cm}p{0.9cm}p{0.9cm}p{0.9cm}p{0.9cm}}
		\toprule   
		{} & \multicolumn{2}{c|}{mAP} & \multicolumn{2}{c}{Car} & \multicolumn{2}{c}{Pedestrian} & \multicolumn{2}{c}{Cyclist} \\
		\cmidrule{2-3}
		\cmidrule{4-5}
		\cmidrule{6-7}
		\cmidrule{8-9}
		Training method & SR & MR & SR & MR & SR & MR & SR & MR \\ 
		\midrule
		$\text{L}^{\text{SSTM}}_{\text{1} }                         $ & 56.3 & 34.0 & 59.8 & 46.3 & 42.2 & 17.5 & 66.7 & 38.3 \\
		\midrule
		$\text{L}^{\text{SSTM}}_{\text{\sfrac{1}{2} / rand} }       $ & 46.2 & 21.2 & 50.8 & 35.4 & 34.1 & \hspace{0.15cm}9.8 & 53.7 & 18.3 \\
		$\text{L}^{\text{SSTM}}_{\text{\sfrac{1}{4} / rand} }       $ & 44.5 & 18.4 & 53.5 & 31.7 & 32.5 & 10.2 & 47.5 & 13.1 \\
		$\text{L}^{\text{SSTM}}_{\text{\sfrac{1}{8} / rand} }       $ & 37.1 & 12.2 & 48.7 & 17.0 & 24.3 & \hspace{0.15cm}9.1 & 38.3 & 10.5 \\
		$\text{L}^{\text{SSTM}}_{\text{\sfrac{1}{16} / rand} }      $ & 28.0 & \hspace{0.15cm}3.1 & 38.6 & \hspace{0.15cm}4.5 & 21.0 & \hspace{0.15cm}1.1 & 24.3 & \hspace{0.15cm}3.7 \\
		$\text{L}^{\text{SSTM}}_{\text{\sfrac{1}{32} / rand} }      $ & 19.7 & \hspace{0.15cm}3.8 & 33.4 & \hspace{0.15cm}9.1 & 10.0 & \hspace{0.15cm}1.5 & 15.5 & \hspace{0.15cm}0.7 \\
		$\text{L}^{\text{SSTM}}_{\text{\sfrac{1}{64} / rand} }      $ & 12.2 & \hspace{0.15cm}1.1 & 21.1 & \hspace{0.15cm}3.0 & \hspace{0.15cm}3.7 & \hspace{0.15cm}0.1 & 11.7 & \hspace{0.15cm}0.1 \\
		$\text{L}^{\text{SSTM}}_{\text{\sfrac{1}{128} / rand} }     $ & 11.2 & \hspace{0.15cm}1.3 & 19.4 & \hspace{0.15cm}3.0 & \hspace{0.15cm}4.9 & \hspace{0.15cm}0.1 & \hspace{0.15cm}9.5 & \hspace{0.15cm}0.8 \\
		$\text{L}^{\text{SSTM}}_{\text{\sfrac{1}{256} / rand} }     $ & 10.8 & \hspace{0.15cm}1.6 & 14.1 & \hspace{0.15cm}4.5 & \hspace{0.15cm}9.1 & \hspace{0.15cm}0.0 & \hspace{0.15cm}9.1 & \hspace{0.15cm}0.2 \\
		\bottomrule
	\end{tabular}
	\label{tab:MSTM3}
\end{table*}

\section{Derivation of KD-losses}

Section 3.1 in the original paper introduces the KD-losses utilized in this work. Three loss terms, as described by \cite{Yang2022KDOpenPCDet} are utilized in this work. This section is supposed to give a more detailed description of all the losses.
\subsection{Logit-KD}
The logit-KD loss $\color{blue}\mathcal{L}_{\text{logit}}$ consists of two sub loss terms, as described by \cite{Yang2022KDOpenPCDet}. First is the bounding box position regression loss:		
\begin{equation}
\label{eqA:Lreg}
\mathcal{L}_{\text{l-reg}} = \mathcal{L}_{\text{reg}}(p_{\text{reg}}^{s}, p_{\text{reg}}^{t}),
\end{equation}

where $p_{\text{reg}}^{s}$ represents the bounding box regression prediction of the student while $p_{\text{reg}}^{t}$ represents the regression prediction of the teacher. $\mathcal{L}_{\text{reg}}$ is the respective regression loss function of the utilized detection algorithm. 

The second loss is the object class loss:
\begin{equation}
\label{eqA:Lcls}
\mathcal{L}_{\text{l-cls}} = \mathbb{E}[ \| \kappa(p^s_{\text{cls}}) - p^t_{\text{cls}} \|_2],
\end{equation}
where $p_{\text{cls}}^{s}$ and $p_{\text{cls}}^{t}$ represent the object classification after the sigmoid activation of the student and teacher network respectively. To match the student classification to the teacher classification the bilinear interpolation $\kappa$ of the student classification is utilized. 

\subsection{Feature-KD} The Feature-KD loss $\color{red}\mathcal{L}_{\text{feature}}$ forces the student to mimic the teacher's intermediate bird's-eye view feature (feat) map:
\begin{equation}
\label{eqA:Lfeat}
\mathcal{L}_{\text{feat}} = \mathbb{E}[\| \psi( \phi(\kappa(f^s)), y) - \psi(f^t, y) \|_2],
\end{equation}
where $f^s$ and $f^t$ represent the student and teacher feature maps respectively, $y$ represents the ground truth labels, $\kappa$ is the bilinear interpolation, which matches the students to the teachers feature map, $\psi$ is the RoI Alignment \cite{he2017mask}, $\phi$ represents a $1 \times 1$ convolution block with batch normalization \cite{Ioffe2015BatchNorm} and ReLU \cite{agarap2018deep} activation to align channel-wise discrepancy between teacher and student.

\subsection{Label-KD}
The Label-KD loss $\mathcal{L}^{*}_{\text{label}}$ replaces the regression and classification loss of any given object detector by constructing a modified ground truth set $\hat{y}^{\text{GT}} = \{y, \hat{y}^t \}$, consisting of the ground truth labels $y$ and the teachers object predictions $\hat{y}^t$, which are filtered by their confidence score by a factor $\tau$.

\section{Evaluation on Voxel R-CNN as an Object Detector}
Table \ref{tab:VoxelRCNN} shows the results of the MSTM and KD on Voxel R-CNN \cite{deng2021voxel} for selected methods. The results coincide with the results on PointPillars. Initializing the student with the teacher's weights as well as the MSTM results in a performance benefit, while the KD approach only contributes to minor improvements.

\begin{table*}
	\centering
	\caption{Radar-only detection performance utilizing Voxel R-CNN as an object detector. The best and second best results are marked in \textbf{bold} and \underline{underlined}, respectively.}
	\begin{tabular}{@{\extracolsep{1pt}}p{3cm}p{0.9cm}p{0.9cm}| p{0.9cm}p{0.9cm}p{0.9cm}p{0.9cm}p{0.9cm}p{0.9cm}}
		\toprule   
		{} & \multicolumn{2}{c|}{mAP} & \multicolumn{2}{c}{Car} & \multicolumn{2}{c}{Pedestrian} & \multicolumn{2}{c}{Cyclist} \\
		\cmidrule{2-3}
		\cmidrule{4-5}
		\cmidrule{6-7}
		\cmidrule{8-9}
		Training method & SR & MR & SR & MR & SR & MR & SR & MR \\ 
		\midrule
		$\text{R}^{\text{SSTM}}                                                 $ & 36.7 & \underline{14.6} & 43.2 & 18.7 & 20.1 & \underline{10.2} & 46.7 & \underline{14.7} \\
		\midrule
		$\text{RL}^{\text{MSTM}}_{\text{1-\sfrac{1}{16} / vox}}\rightarrow\text{R}                         $ & 37.7 & \textbf{15.1} & \textbf{44.6} & \underline{19.6} & 22.3 & \textbf{10.4} & 46.2 & \textbf{15.2} \\
		\midrule
		$\text{RL}^{\text{SSTM}}_{\text{\sfrac{1}{4} / vox}}\rightarrow\text{R}                  $ & \underline{38.5} & 14.5 & 44.1 & \textbf{20.5} & \underline{22.9} & \hspace{0.15cm}9.1 & \underline{48.5} & 13.9 \\
		$\text{RL}^{\text{SSTM}}_{\text{\sfrac{1}{4} / vox}}\rightarrow\text{R}^{\text{feat}}     $ & \textbf{39.0} & 14.1 & \underline{44.3} & 19.3 & \textbf{23.3} & \hspace{0.15cm}9.3 & \textbf{49.3} & 13.8 \\
		\bottomrule
	\end{tabular}
	\label{tab:VoxelRCNN}
\end{table*}

\section{Hyperparameters}
For PointPillars, we adopted the configuration from the VoD dataset \cite{palffy2022voddataset}; for Voxel R-CNN, we used the standard setup utilized on the Kitti Dataset in OpenPCDet \cite{openpcdet2020}; for DSVT-P we used the standard setup utilized on the Waymo Dataset with the changes described in our main paper and the hyperparameters listed in Table \ref{tab:params_poinpillars} - \ref{tab:params_kd}. Further details on parameter configurations see the model configurations in OpenPCDet \cite{openpcdet2020} or our code release: \url{https://github.com/rst-tu-dortmund/lerojd}.

\begin{table}[h]
	\centering
	\caption{Parameters used for PointPillars.}
	\begin{tabular}{c|c}
		\toprule
		Parameter & Value  \\ \midrule
		Voxel size & $0.16 \, \text{m} \times 0.16  \, \text{m} \times 5  \, \text{m} $ \\
		Max \#points/pillar & 32 \\
		Point cloud range - $x$ & [0 m, 51.2 m] \\
		Point cloud range - $y$ & [-25.6 m, 25.6 m] \\
		Point cloud range - $z$ & [-3 m, 2 m] \\
		Learning rate & 0.003 \\
		\bottomrule
	\end{tabular}
	\label{tab:params_poinpillars}
\end{table}

\begin{table}[ht]
	\centering
	\caption{Parameters used for Voxel R-CNN.}
	\begin{tabular}{c|c}
		\toprule
		Parameter & Value  \\ \midrule
		Voxel size & $0.036 \, \text{m} \times 0.032  \, \text{m} \times 0.125  \, \text{m} $ \\
		Max \#points/voxel & 32 \\
		Point cloud range - $x$ & [0 m, 51.2 m] \\
		Point cloud range - $y$ & [-25.6 m, 25.6 m] \\
		Point cloud range - $z$ & [-3 m, 2 m] \\
		Learning rate & 0.01 \\
		\bottomrule
	\end{tabular}
	\label{tab:params_voxelrcnn}
\end{table}

\begin{table}[ht]
	\centering
	\caption{Parameters used for DSVT-P.}
	\begin{tabular}{c|c}
		\toprule
		Parameter & Value  \\ \midrule
		Voxel size & $0.2031 \, \text{m} \times 0.2031  \, \text{m} \times 5  \, \text{m} $ \\
		Max \#points/pillar & 32 \\
		Point cloud range - $x$ & [0 m, 51.2 m] \\
		Point cloud range - $y$ & [-25.6 m, 25.6 m] \\
		Point cloud range - $z$ & [-3 m, 2 m] \\
		Learning rate & 0.0003 \\
		Sparse shape & [252, 252, 1] \\
		Window size & [12, 12, 1] \\
		Hybrid factor & [2, 2, 1] \\
		Input dimension & [[0, 0, 0], [6, 6, 0]] \\
		\# Layers & 4\\
		NMS threshold & 0.01\\
		\bottomrule
	\end{tabular}
	\label{tab:params_dsvt}
\end{table}

\begin{table}[ht]
	\centering
	\caption{Parameters used for Voxel-based sampling.}
	\begin{tabular}{c|c}
		\toprule
		Parameter & Value  \\ \midrule
		Voxel size & $1 \, \text{m} \times1  \, \text{m} \times1  \, \text{m} $ \\
		\bottomrule
	\end{tabular}
	\label{tab:params_voxelsampling}
\end{table}

\begin{table}[ht]
	\centering
	\caption{Weighting of knowledge distillation losses.}
	\begin{tabular}{c|c}
		\toprule
		Parameter & Value  \\ \midrule
		$\lambda_{\text{reg}}$ & 1.0 \\
		$\lambda_{\text{cls}}$ & 1.0 \\
		$\lambda_{\text{feat}}$ & 0.1 \\
		$\lambda_{\text{l-reg}}$ & 0.3 \\
		$\lambda_{\text{l-cls}}$ & 0.001 
		\\ \bottomrule
	\end{tabular}
	\label{tab:params_kd}
\end{table}





%
%

\end{document}